\documentclass[10pt,twocolumn,letterpaper]{article}

\usepackage{iccv,times,epsfig,graphicx,amsmath,amssymb}
\usepackage{booktabs,stfloats,xcolor,colortbl,multirow,pifont}
\usepackage[export]{adjustbox} 

\usepackage[pagebackref=false, breaklinks=true, letterpaper=true, colorlinks,
            citecolor=citecolor, linkcolor=linkcolor, bookmarks=false]{hyperref}
\definecolor{citecolor}{HTML}{0071BC}
\definecolor{linkcolor}{HTML}{ED1C24}

\definecolor{MyDarkBlue}{rgb}{0,0.08,1}
\definecolor{MyDarkGreen}{rgb}{0.02,0.6,0.02}
\definecolor{MyDarkRed}{rgb}{0.8,0.02,0.02}
\definecolor{MyDarkOrange}{rgb}{0.40,0.2,0.02}
\definecolor{MyPurple}{RGB}{111,0,255}
\definecolor{MyRed}{rgb}{1.0,0.0,0.0}
\definecolor{MyGold}{rgb}{0.75,0.6,0.12}
\definecolor{MyDarkgray}{rgb}{0.66, 0.66, 0.66}
\definecolor{MyDarkCyan}{rgb}{0.05, 0.55, 0.45}
\definecolor{MyBlack}{rgb}{0., 0., 0.}
\definecolor{MyMagenta}{rgb}{1., 0., 1.}
\definecolor{BerkeleyYellow}{RGB}{255,204,41}
\definecolor{BerkeleyLightBlue}{RGB}{94,146,221}
\definecolor{BkDarkBlue}{rgb}{.05,.07,.353}

\definecolor{color4}{rgb}{0.94,0.94,1}

\newcommand{\MYhref}[3][blue]{\href{#2}{\color{#1}{#3}}} 

\renewcommand{\paragraph}[1]{\vspace{1.25mm}\noindent\textbf{#1}}

\newcommand{\cmark}{\ding{51}}%
\newcolumntype{x}[1]{>{\centering\arraybackslash}p{#1pt}}

\iccvfinalcopy 

\def\iccvPaperID{****} 

\ificcvfinal\fi

\begin{document}

\title{Masked Autoencoders as Image Processors}

\author{Huiyu~Duan$^{1,2}$,~Wei~Shen$^{2}$,~Xiongkuo~Min$^{1}$,~Danyang~Tu$^{1,2}$,~Long~Teng$^{1,2}$,~Jia~Wang$^{1}$,~Guangtao~Zhai$^{1,2}$\\
$^1$Institute of Image Communication and Network Engineering, Shanghai Jiao Tong University\\
$^2$MoE Key Lab of Artificial Intelligence, AI Institute, Shanghai Jiao Tong University\\
{\tt\small \{huiyuduan,~wei.shen,~minxiongkuo,~danyangtu,~tenglong,~jiawang,~zhaiguangtao\}@sjtu.edu.cn}
}

\maketitle
\ificcvfinal\thispagestyle{empty}\fi


\begin{abstract}
Transformers have shown significant effectiveness for various vision tasks including both high-level vision and low-level vision.
Recently, masked autoencoders (MAE) for feature pre-training have further unleashed the potential of Transformers, leading to state-of-the-art performances on various high-level vision tasks.
However, the significance of MAE pre-training on low-level vision tasks has not been sufficiently explored.
In this paper, we show that masked autoencoders are also scalable self-supervised learners for image processing tasks.
We first present an efficient Transformer model considering both channel attention and shifted-window-based self-attention termed CSformer.
Then we develop an effective MAE architecture for image processing (MAEIP) tasks.
Extensive experimental results show that with the help of MAEIP pre-training, our proposed CSformer achieves state-of-the-art performance on various image processing tasks, including Gaussian denoising, real image denoising, single-image motion deblurring, defocus deblurring, and image deraining.
The code and models will be available at: \MYhref[MyMagenta]{https://github.com/DuanHuiyu/MAEIP_CSformer}{https://github.com/DuanHuiyu/MAEIP\_CSformer}.
\end{abstract}

\section{Introduction}
\label{sec:intro}

Image processing, including image restoration, image enhancement, \textit{etc.} has long been an important computer vision task, which aims at improving the image quality of a degraded input.
A good image processor can not only produce images that are more conformed to human visual preferences, but also improve the performance of downstream computer vision tasks such as recognition, detection, segmentation, \textit{etc} \cite{mspfn2020}.
Due to the ill-posed nature of this problem, it generally requires strong image priors for effective processing.
Recently, deep learning has been widely employed to image processing tasks, such as deraining, deblurring, denoising, enhancement, \textit{etc.}, and has achieved leading performance due to its strong ability to learn generalizable priors from large-scale data \cite{ulyanov2018dip,chen2021IPT,zamir2022restormer}.

\begin{figure}[t]\centering
\includegraphics[width=0.97\linewidth]{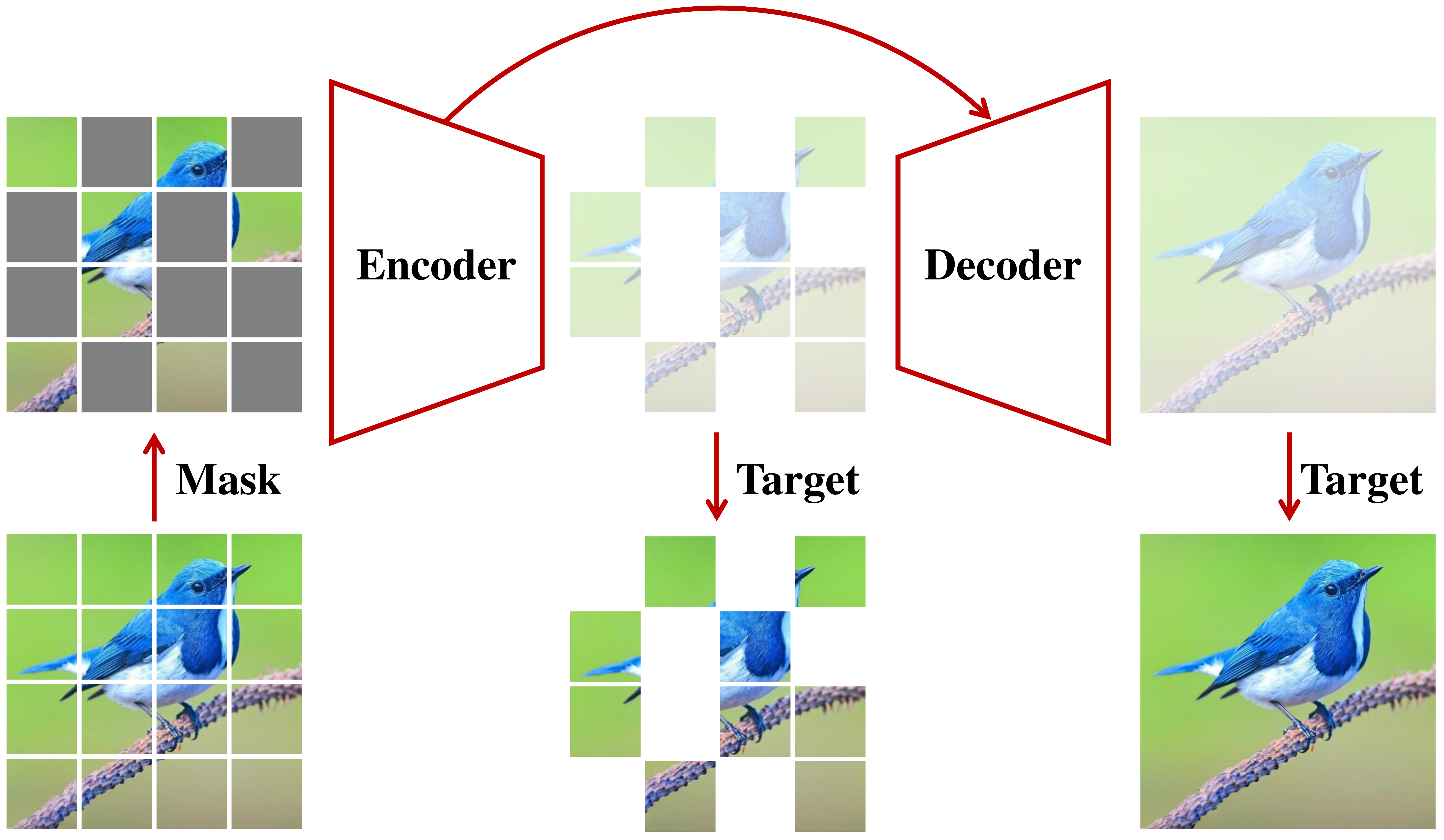}
\caption{An illustration of the proposed MAEIP framework. The encoder predicts the raw pixel values of the randomly masked patches while the decoder predicts the raw pixel values of the whole image.
}
\vspace{-1em}
\label{fig:maeip}
\end{figure}

Convolutional neural networks (CNNs) have been adopted in various low-level vision tasks for many years \cite{DnCNN,zhang2017learning,gopro2017,zhang2020rdn,ren2019progressive,mspfn2020,duan2022develop}, and have achieved impressive results.
However, CNNs generally suffer from the limitation of capturing long-range pixel dependencies.
Recent studies on Vision Transformers (ViT) \cite{carion2020detr,dosovitskiy2021vit,liu2021swin} have explored their potential as alternatives to CNNs considering their effectiveness in capturing long-range dependencies.
Some studies have also explored to use Transformers in low-level vision tasks \cite{chen2021IPT,liang2021swinir,wang2021uformer,zamir2022restormer,tu2022maxim}, and also demonstrated the superiority of the architecture.

Besides the improvement of architectural design, recent self-supervised learning frameworks, such as DINO \cite{caron2021dino}, MOCO-V3 \cite{chen2021mocov3}, MAE \cite{he2022mae}, have further unleashed the potential of ViT and achieved high performance on various high-level vision tasks \cite{he2016resnet,he2017maskrcnn}.
Among them, masked autoencoders (MAE) \cite{bao2022beit,zhou2022ibot,he2022mae,xie2022simmim,gao2022convmae}, which pre-train image models by predicting masked tokens from seen tokens, have demonstrated superior learning ability and scalability on various high-level vision tasks.
However, few studies have generalized the self-supervised pre-training strategy to image processing tasks \cite{chen2021IPT}.

\begin{figure*}[t]\centering
\includegraphics[width=0.98\linewidth]{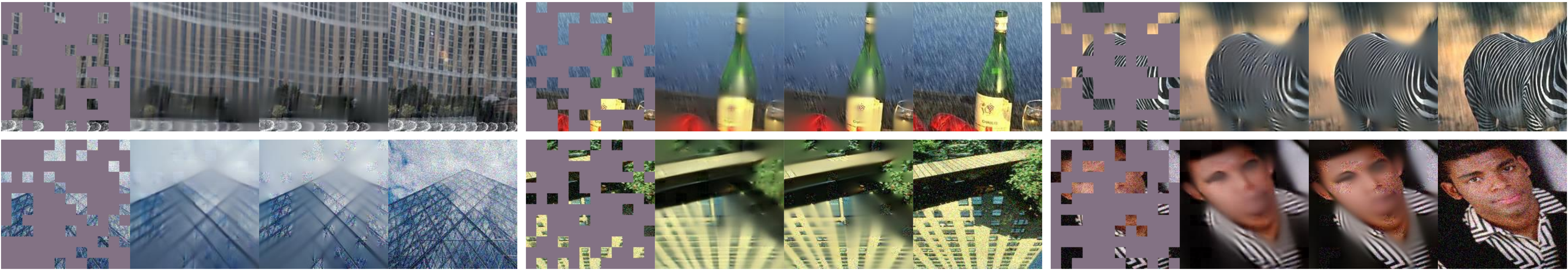}
\caption{Example results on rainy and noisy images. For each quadruplet, we demonstrate the masked image, the reconstruction result from the encoder, the reconstruction result from the decoder, and the raw degraded image, from the left to the right, respectively.
}
\label{fig:maeip_results}
\vspace{-1.2em}
\end{figure*}

In this paper, we show that masked autoencoders are also scalable self-supervised learners for image processing tasks.
We first present an efficient Transformer for image processing.
Recent studies on image processing have demonstrated the superiority of hierarchical structure \cite{wang2021uformer}, window-based self-attention \cite{liang2021swinir}, or channel-wise attention \cite{zamir2022restormer,chen2022nafnet}, in terms of both performance and efficiency.
Our proposed new architecture considers both channel attention and shifted-window-based self-attention and is built based on the hierarchical U-shape structure \cite{ronneberger2015unet}, which is named as CSformer.
Based on the highly efficient CSformer, we further present a MAE pre-training framework for image processing (MAEIP).
As shown in Figure \ref{fig:maeip}, Our MAEIP masks random patches from the input image and reconstructs the masked patches and the whole image by the encoder and the decoder respectively.
Similar to SIMMIM \cite{xie2022simmim}, the masked input is directly fed into the encoder.
Moreover, we adpot the U-shape structure \cite{ronneberger2015unet} for high-quality whole image reconstruction.

\begin{figure}[!t]
\begin{center}
\begin{tabular}{c c}\hspace{-1.14em}
\includegraphics[width=0.2432\textwidth]{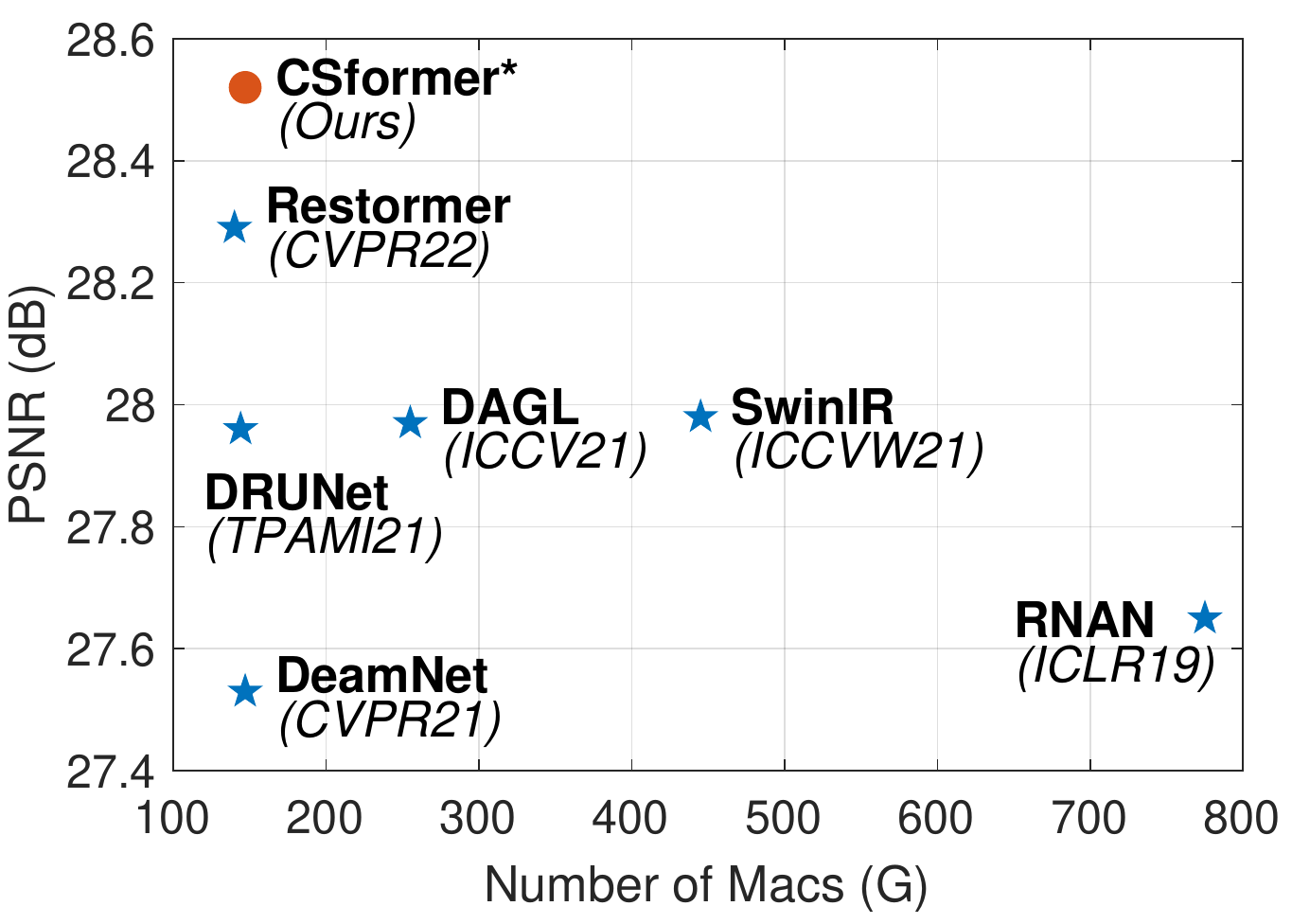}
& \hspace{-1.63em}
\includegraphics[width=0.2432\textwidth]{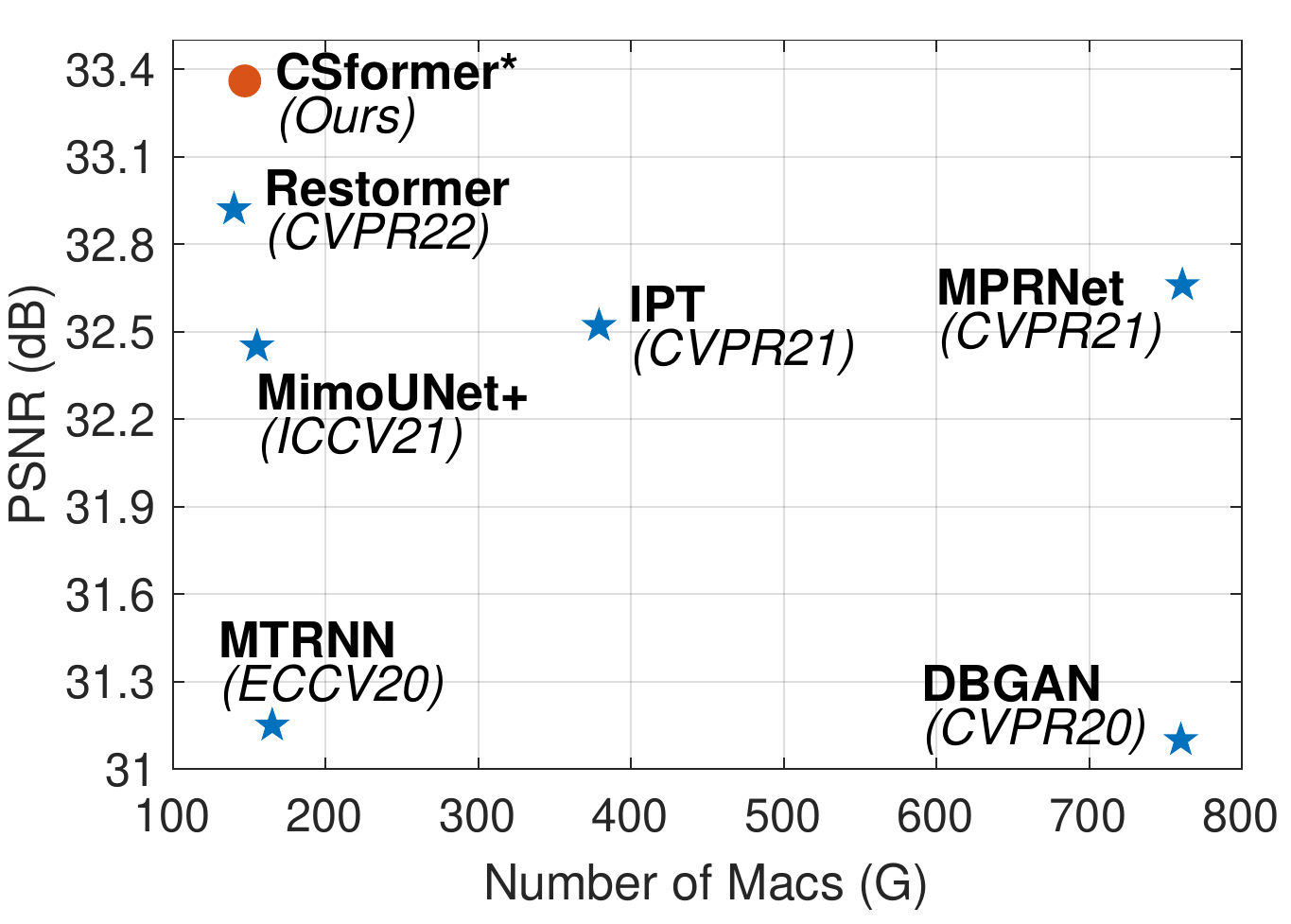}
\vspace{-0.3em}
\\
 \hspace{-0.59em}\small~(a)~Gaussian~Denoising~(Table \ref{table:graydenoising})
&  \hspace{-1.14em}
\small~(b)~Deblurring~(Table \ref{table:deblurring_motion}) 
\\
\end{tabular}
\end{center}
\vspace*{-0.71em}
\caption{PSNR results \textit{vs.} computational cost.
}
\label{fig:performance}
\vspace{-1.2em}
\end{figure}

Figure \ref{fig:maeip_results} gives several example results of our MAEIP framework on masked distorted images.
Both the reconstruction results from the encoder and the decoder on masked pixels show that the MAE pre-training can well resist noise from the seen input and well reconstruct the expected image, which is a natural image processor.
Moreover, the decoder can well keep the detailed information from the seen input, which leads to a higher-quality and more consistent reconstruction of the whole image.
With the help of this effective pre-training strategy, our CSformer achieves state-of-the-art (SOTA) performance on various benchmark datasets for different image processing tasks as shown in Figure \ref{fig:performance}.

\section{Related Work}\label{sec:related}

\subsection{Image Processing}

With the development of deep learning methods and the establishment of various vision benchmarks, data-driven CNN architectures \cite{DnCNN,zamir2020mirnet,Zamir_2021_CVPR_mprnet,zhang2018image,zhang2020rdn} have attained state-of-the-art performance on various image processing tasks compared to conventional processing approaches \cite{BM3D,he2010single,timofte2013anchored}.
The architecture design plays an important role in improving performance, and many studies have developed general-purpose or task-specific modules for various image processing applications.
Encoder-decoder-based U-Net architectures \cite{ronneberger2015unet} have been widely adopted for image processing due to their high computational efficiency.
Moreover, many advanced components developed for high-level vision tasks have also been introduced to low-level vision tasks and demonstrated the effectiveness, such as residual and dense connection \cite{DnCNN,ledig2017photo,wang2018esrgan,zhang2020rdn}, channel attention \cite{niu2020single,zamir2020mirnet,Zamir_2021_CVPR_mprnet,duan2022develop}, spatial attention \cite{wang2018non,liu2018non,Zamir_2021_CVPR_mprnet,duan2022develop}, multi-scale or multi-stage networks \cite{dmphn2019,deblurganv2,mspfn2020,cho2021rethinking_mimo,Zamir_2021_CVPR_mprnet,chen2021hinet}, \textit{etc.}

Recently, Transformers, which are originally developed for NLP tasks \cite{vaswani2017attention}, have been introduced to computer vision and achieved state-of-the-art performance \cite{carion2020detr,dosovitskiy2021vit}.
Many transformer-based architectures for high-level vision tasks have also been proposed such as ViT \cite{dosovitskiy2021vit}, Swin \cite{liu2021swin}, PVT \cite{wang2021pyramid}, \textit{etc}.
Several recent studies have also explored specific Transformer architectures for low-level vision problems.
Based on DETR \cite{carion2020detr} and ViT \cite{dosovitskiy2021vit}, IPT \cite{chen2021IPT} applies the standard Transformer encoder-decoder architecture \cite{vaswani2017attention} to image patches and presents a pre-training method for low-level vision tasks.
However, IPT relies on large-scale datasets and multi-task learning for good performance and has huge computing complexity \cite{liang2021swinir,zamir2022restormer}.
SwinIR \cite{liang2021swinir} and Uformer \cite{wang2021uformer} adopt the shifted-window-based local attention module \cite{liu2021swin} in low-level tasks.
However, restricting the spatial attention to local windows may also limit the long-range dependency capturing of the whole image \cite{zamir2022restormer}.
Restormer \cite{zamir2022restormer} presents a multi-dconv head transposed attention (MDTA) block, which applies spatial attention across feature dimensions to reduce computational complexity.
This operation can implicitly learn spatial correlation, which could be seen as a special variant of channel attention \cite{chen2022nafnet}.

\vspace{-0.1em}
\subsection{MAE Pre-training}
\vspace{-0.1em}

Masked language modeling approaches such as BERT \cite{kenton2019bert} and GPT \cite{radford2018improving,radford2019language,brown2020language} are successful methods for pre-training NLP models.
Recently, some works have also explored the masking strategy in pre-training image models.
iGPT \cite{chen2020igpt} considers image pixels as sequences and predicts unknown pixels.
BEiT \cite{bao2022beit} and iBOT \cite{zhou2022ibot} propose to predict discrete tokens rather than pixels to pre-train image models.
MAE \cite{he2022mae} finds that masking a high proportion of the input image can contribute to meaningful self-supervised learning, and proposes an asymmetric encoder-decoder structure to reduce pre-training time.
SimMIM \cite{xie2022simmim} and ConvMAE \cite{gao2022convmae} propose general masked image modeling methods for hierarchical models such as Swin Transformer \cite{liu2021swin} and hybrid convolution-transformer \cite{gao2022convmae}.
These pre-training methods are designed for high-level vision tasks, while few studies explore self-supervised pre-training methods for low-level vision applications.

\begin{figure*}[t]\centering
\vspace{-0.5em}
\includegraphics[width=0.98\linewidth]{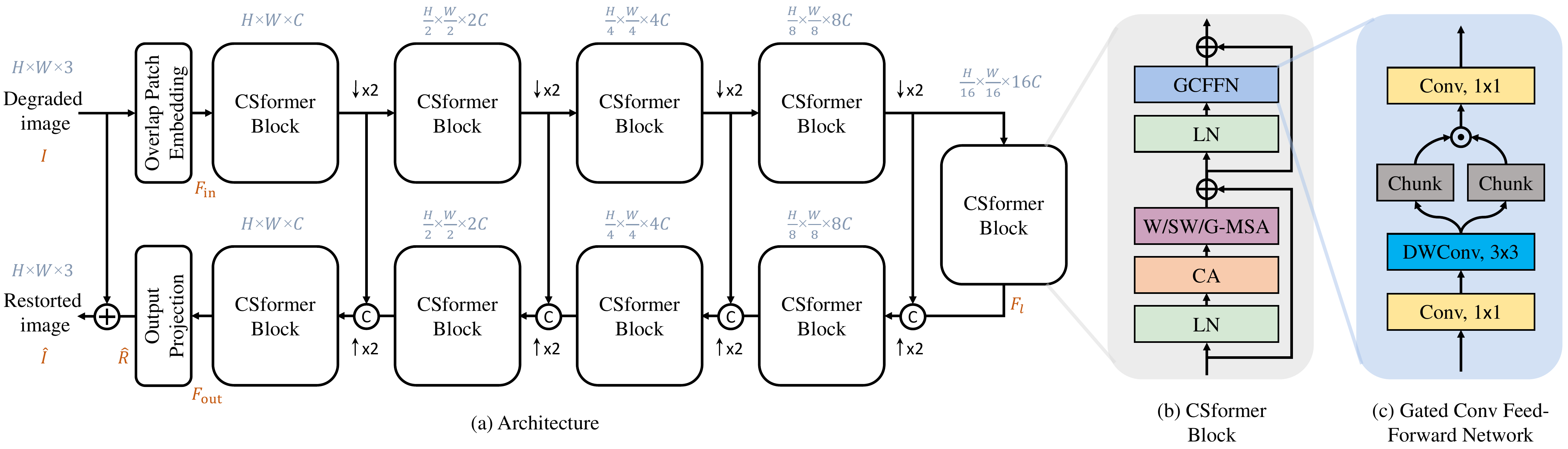}
\vspace{-0.2em}
\caption{An overview of the proposed CSformer. (a) The architecture of the CSformer backbone. (b) An illustration of a CSformer block. (c) An illustration of the Gated Conv Feed-Forward Network (GCFFN). LN indicates a LayerNorm layer. CA is a channel-attention layer. W/SW/G-MSA represent multi-head self-attention within the regular window, shifted window, and global image, respectively. 
}
\label{fig:csformer}
\vspace{-1.3em}
\end{figure*}


\vspace{-0.5em}
\section{Approach}
\label{sec:approach}
\vspace{-0.2em}

In this section, we first introduce the detailed architecture of the proposed CSformer, then we provide the details of our framework of masked autoencoders for image processing (MAEIP).

\vspace{-0.1em}
\subsection{CSformer}
\vspace{-0.2em}
\paragraph{Overall pipeline.} Figure \ref{fig:csformer}-(a) demonstrates the architecture of the CSformer backbone. Our CSformer follows the design principles of encoder-decoder with skip connections, which is originally proposed by UNet \cite{ronneberger2015unet}. Given an input degraded image $\textbf{I} \in \mathbb{R}^{H \times W \times 3}$, CSformer first applies a $3 \times 3$ convolutional layer to extract low-level feature embeddings $\textbf{F}_\text{in} \in \mathbb{R}^{H \times W \times C}$.
Next, these shallow feature maps $\textbf{F}_\text{in}$ are passed through a 5-level symmetric encoder-decoder network, then output feature maps $\textbf{F}_\text{out} \in \mathbb{R}^{H \times W \times C}$. Each stage of the encoder-decoder contains multiple CSformer blocks, and the bottleneck stage outputs latent features $\textbf{F}_l \in \mathbb{R}^{\frac{H}{16} \times \frac{W}{16} \times 16C}$.
The specific feature dimensions of each stage are shown in Figure \ref{fig:csformer}-(a).
We apply pixel-unshuffle and pixel-shuffle \cite{shi2016real} for the down-sampling process in the encoder and the up-sampling process in the decoder, respectively.
Similar to UNet \cite{ronneberger2015unet}, we add skip connections between the corresponding stages of the encoder and decoder, and first concatenate two feature maps then process the concatenated features with a $1 \times 1$ convolution to reduce the channel dimension.
The output feature maps $\textbf{F}_\text{out}$ from the encoder-decoder network are finally refined with a $3 \times 3$ convolutional layer to get the estimated residual map $\hat{\textbf{R}} \in \mathbb{R}^{H \times W \times 3}$.
The final restored image is obtained by $\hat{\textbf{I}} = \textbf{I} + \hat{\textbf{R}}$.
We optimize the CSformer using the Charbonnier loss \cite{charbonnier1994two,liang2021swinir}: {\small $\mathcal{L}= \sqrt{(|| \hat{\textbf{I}}-\textbf{I}' ||)^2 + \varepsilon^2}$, where $\textbf{I}'$} is the ground-truth image, and $\varepsilon=10^{-3}$ is a constant.

\paragraph{CSformer block.} Applying self-attention among global image pixels can lead to huge computational costs \cite{chen2020igpt,liu2021swin,chen2021IPT}. However, image processing problems need to consider both local context information and global context information \cite{liang2021swinir,wang2021uformer,zamir2022restormer}, while high-level vision transformers such as ViT show limitations in capturing local pixel dependencies \cite{li2021localvit,wang2021uformer}.
As shown in Figure \ref{fig:csformer}-(b), similar to the standard transformer architecture \cite{vaswani2017attention,dosovitskiy2021vit}, a CSformer block consists of two parts with skip connections including an attention component and a feed-forward part.
The attention component contains a channel-attention (CA) module and a window-based/shifted-window-based/global multi-head self-attention (W/SW/G-MSA) with applying a LayerNorm (LN) layer before.
We apply W/SW/G-MSA across different stages.
Specifically, for the bottleneck stage, whose channel-dimension is $16C$, we apply the global MSA to aggregate global context information, while for other stages, of which the channel-dimensions are from $C$ to $8C$, we alternately apply W-MSA and SW-MSA respectively, which is similar to \cite{liu2021swin}.
We adopt the gated convolutional feed-forward network (GCFFN) \cite{dauphin2017language,shazeer2020glu,zamir2022restormer} as the feed-forward component with a LN layer before.

\paragraph{Channel-attention.} We adopt a simplified channel-attention module \cite{chen2022nafnet} rather than the raw squeeze-and-excitation (SE) module \cite{hu2018squeeze}.
Given an input tensor $\textbf{X}$, the output of the CA layer can be formulated as: $\text{CA}(\textbf{X})=\textbf{X}*\text{MLP}(\text{Avg}(\textbf{X}))$, where Avg is an average pooling layer, MLP is a multilayer perceptron, $*$ indicates a channel-wise product operation.
Before the channel-attention layer, a simple gate network \cite{dauphin2017language,shazeer2020glu,chen2022nafnet} is also applied to control which complementary features should be adopted.

\paragraph{Gated convolutional feed-forward network.} Conventional FFN \cite{vaswani2017attention,li2021localvit} adopts two multilayer perceptrons (MLPs) to first expand then reduce the channel dimensions to refine features.
Some studies on low-level vision have manifested that adopting a lightweight depth-wise convolution in FFN can effectively enhance the locality and further improve the output quality \cite{wang2021uformer,zamir2022restormer}.
As shown in Figure \ref{fig:csformer}-(c), our CSformer adopts a gated convolutional feed-forward network (GCFFN) \cite{dauphin2017language,shazeer2020glu,zamir2022restormer}.
Give an input tensor $\textbf{X}$, the GCFFN process can be formulated as: $\textbf{X}_1 = W_d^1(W_p^1(\textbf{X}))$, $\textbf{X}_2 = W_d^2(W_p^2(\textbf{X}))$, $\hat{\textbf{X}} = W_p^3(\phi(\textbf{X}_1)\odot \textbf{X}_2)$, where $W_p$ represents a $1 \times 1$ point-wise convolution, $W_d$ indicates a $3 \times 3$ depth-wise convolution, $\phi$ is the GELU operation, $\odot$ denotes element-wise multiplication.

\paragraph{Inference process speed-up.} The Swin structure \cite{liu2021swin} is hard to test on arbitrary input size due to the hierarchical window limitation.
Previous works generally need to pad the input image to a larger size as a new input, then feed it into the network \cite{wang2021uformer}.
This padding strategy can cause huge extra computational overhead.
For instance, the window size in Uformer \cite{wang2021uformer} is set to 8, which means the bottleneck layer needs to be padded to a multiple of 8.
Since there are five hierarchical stages in Uformer, the input needs to be padded to a multiple of 128 during the inference process.
We propose to pad the feature rather than pad the input to save extra computational overhead.
Specifically, for each layer, we pad its size to the multiple of 8 and use a mask to prevent the attention calculation on the padded area.
Given an extreme example, for a $16 \times 16$ image, the spatial dimensions of the hierarchical stages in Uformer are
$128 \times 128$, $64 \times 64$, $32 \times 32$, $16 \times 16$, $8 \times 8$, respectively, while in our CSformer are $16 \times 16$, $8 \times 8$, $8 \times 8$, $8 \times 8$, $1 \times 1$ (this layer is not a window-based layer), respectively.

\subsection{MAEIP}
Similar to MAE \cite{he2022mae} and SIMMIM \cite{xie2022simmim}, our MAEIP is a simple autoencoding approach, which masks a portion of image signals and learns to reconstruct them.
We implement the MAEIP pre-training based on the CSformer.
We expect that the encoder can learn more meaningful image representations, while the decoder can exploit the image representations from the encoder and the low-level details from the skip connections to reconstruct higher-quality images.
Figure \ref{fig:maeip} illustrates the idea and detailed methods are introduced as follows.

\paragraph{Masking.} 
Masked autoencoders such as BEiT \cite{bao2022beit}, MAE \cite{he2022mae}, and SimMIM \cite{xie2022simmim}, first embed image patches to tokens, and then adopt a random mask on input tokens.
However, the same strategy cannot be directly used in our MAEIP, since low-level vision transformers including our CSformer generally adopt the overlap patch embedding strategy.
Therefore, in our MAEIP, we directly mask random image patches on raw image pixels.
Then the masked image is fed into the autoencoder to learn to reconstruct the original image.
We follow the conventions in \cite{he2022mae,xie2022simmim} and mask random patches with $16 \times 16$ pixels, and adopt a high masking ratio \textit{i.e.,} 75\%.

\paragraph{MAEIP encoder.} 
The encoder in CSformer is defined as from the masked input image $\textbf{I}$ to the output of the bottleneck stage, \textit{i.e.,} $\textbf{F}_l$.
Just as the normal training process of CSformer, the encoder directly embeds the masked images with a $3 \times 3$ convolutional layer, and then processes the overlap embedding via a series of hierarchical CSformer blocks to get the latent features $\textbf{F}_l$.
The encoder reconstructs the masked input by predicting the pixel values for each masked patch.
Specifically, we apply a linear layer to project the latent features $\textbf{F}_l$ to patch pixels \cite{xie2022simmim}, and compute the mean squared error (MSE) between the reconstructed and original images on the masked pixels \cite{he2022mae}.

\paragraph{MAEIP decoder.}
The decoder in CSformer is defined as the process from $\textbf{F}_l$ to the reconstructed image $\hat{\textbf{I}}$.
As shown in Figure \ref{fig:maeip_results}, the reconstructed results from the encoder can only keep high-level image representations while losing image details, and may exist block artifacts between the masked area and seen area.
To reconstruct higher-quality images and keep image details for facilitating subsequent image processing tasks, the MAEIP decoder learns to reconstruct the whole image.
We remove the skip connection between $\textbf{I}$ and $\hat{\textbf{I}}$ in CSformer during the MAEIP pre-training, which may affect the optimization target.
The decoder reconstructs the whole image by directly calculating the MSE loss between the reconstructed image $\hat{\textbf{I}}$ and the original image $\textbf{I}$ on all pixels.

\paragraph{Two-stage pre-training.}
Since our CSformer is a symmetric encoder-decoder network, pre-training the whole network may introduce huge computational overhead.
Learning image representation is a more difficult task \cite{he2022mae}.
Therefore, we propose to pre-train MAEIP using a two-stage paradigm.
In the first stage, we discard the decoder and only pre-train the MAEIP encoder.
For the second stage, we apply the pre-trained encoder weights and pre-train the whole autoencoder.
Experiments in Sec. \ref{sec:ablation} demonstrate this two-stage pre-training scheme achieves similar performance compared with the one-stage pre-training while saving pre-training time.

\begin{table*}[t]
\parbox{.435\linewidth}{
\centering
\caption{\textbf{Grayscale image denoising} results of different models. Top super row (blind denoising): learning a single model and testing on various noise levels. Bottom super row (non-blind denoising): training a separate model for each noise level.} 
\label{table:graydenoising}
\vspace{0.42em}
\setlength{\tabcolsep}{0.15em}
\scalebox{0.7}{
\begin{tabular}{l | c c c | c c c | c c c}
\toprule[0.15em]
   & \multicolumn{3}{c|}{\textbf{Set12}~\cite{DnCNN}} & \multicolumn{3}{c|}{\textbf{BSD68}~\cite{martin2001database_bsd}} & \multicolumn{3}{c}{\textbf{Urban100}~\cite{huang2015single_urban100}} \\
 \cline{2-10}
   \textbf{Method} & $\sigma$$=$$15$ & $\sigma$$=$$25$ & $\sigma$$=$$50$ & $\sigma$$=$$15$ & $\sigma$$=$$25$ & $\sigma$$=$$50$ & $\sigma$$=$$15$ & $\sigma$$=$$25$ & $\sigma$$=$$50$ \\
\midrule[0.15em]
DnCNN~\cite{DnCNN}  &32.67 & 30.35 & 27.18 & 31.62 & 29.16 & 26.23 & 32.28 & 29.80 & 26.35\\
FFDNet~\cite{FFDNetPlus}  &32.75 & 30.43 & 27.32 & 31.63 & 29.19 & 26.29 & 32.40 & 29.90 & 26.50\\ 
IRCNN~\cite{zhang2017learning}  &32.76 & 30.37 & 27.12 & 31.63 & 29.15 & 26.19 & 32.46 & 29.80 & 26.22\\ 
DRUNet~\cite{zhang2021DPIR}  & 33.25 & 30.94 & 27.90 & \underline{31.91} & \underline{29.48} & 26.59 & 33.44 & 31.11 & 27.96\\ 
Restormer \cite{zamir2022restormer} & \underline{33.35}	& \underline{31.04}	& \underline{28.01} & \textbf{31.95}	& \textbf{29.51}	& \textbf{26.62} & \underline{33.67}	& \underline{31.39}	& \underline{28.33}\\ 
 \textbf{CSformer}$^*$ & \textbf{33.40} & \textbf{31.09} & \textbf{28.06} & \textbf{31.95} & \textbf{29.51}& \underline{26.60}& \textbf{33.74} & \textbf{31.51}& \textbf{28.49}\\ 
\midrule[0.1em]
\midrule[0.1em]
FOCNet~\cite{jia2019focnet}  &33.07 & 30.73 & 27.68 & 31.83 & 29.38 & 26.50 & 33.15 & 30.64 & 27.40\\
MWCNN~\cite{liu2018MWCNN}  &33.15 & 30.79 & 27.74 & 31.86 & 29.41 & 26.53 & 33.17 & 30.66 & 27.42\\
NLRN~\cite{liu2018NLRN}  &33.16 & 30.80 & 27.64 & 31.88 & 29.41 & 26.47 & 33.45 & 30.94 & 27.49\\
RNAN~\cite{zhang2019residual}  &- & - & 27.70 & - & - & 26.48 & - & - & 27.65\\
DeamNet~\cite{ren2021adaptivedeamnet}  &33.19 & 30.81 & 27.74 & 31.91 & 29.44 & 26.54 & 33.37 & 30.85 & 27.53 \\
DAGL~\cite{mou2021dynamicDAGL}  &33.28 & 30.93 & 27.81 & 31.93 & 29.46 & 26.51 & 33.79 & 31.39 & 27.97 \\
SwinIR~\cite{liang2021swinir} & 33.36 & 31.01 & 27.91 & \textbf{31.97} & \underline{29.50} & 26.58 & 33.70 & 31.30 & 27.98 \\ 
Restormer \cite{zamir2022restormer} & \underline{33.42} & \underline{31.08} & \underline{28.00} & \underline{31.96} & \textbf{29.52}& \textbf{26.62}& \underline{33.79} & \underline{31.46}& \underline{28.29}\\ 
 \textbf{CSformer}$^*$ & \textbf{33.44} & \textbf{31.10} & \textbf{28.08} & \textbf{31.97} & \textbf{29.52}& \underline{26.60}& \textbf{33.83} & \textbf{31.52}& \textbf{28.52}\\ 
\bottomrule[0.1em]
\end{tabular}}
}
\hfill
\parbox{.54\linewidth}{
\centering
\caption{\textbf{Color image denoising} results of different models. Our CSformer$^*$ demonstrates great performance for both blind and non-blind denoising. On Urban dataset \cite{huang2015single_urban100} with a noise level of 50, our CSformer$^*$ achieves 0.15 dB gain and 0.17 dB gain compared to Restormer \cite{zamir2022restormer} for blind condition and non-blind condition, respectively.
}
\label{table:colordenoising}
\vspace{0.23em}
\setlength{\tabcolsep}{0.15em}
\scalebox{0.7}{
\begin{tabular}{l | c c c | c c c | c c c | c c c}
\toprule[0.15em]
   & \multicolumn{3}{c|}{\textbf{CBSD68}~\cite{martin2001database_bsd}} & \multicolumn{3}{c|}{\textbf{Kodak24}~\cite{kodak}} & \multicolumn{3}{c|}{\textbf{McMaster}~\cite{zhang2011color_mcmaster}} & \multicolumn{3}{c}{\textbf{Urban100}~\cite{huang2015single_urban100}} \\
 \cline{2-13}
   \textbf{Method} & $\sigma$$=$$15$ & $\sigma$$=$$25$ & $\sigma$$=$$50$ & $\sigma$$=$$15$ & $\sigma$$=$$25$ & $\sigma$$=$$50$ & $\sigma$$=$$15$ & $\sigma$$=$$25$ & $\sigma$$=$$50$ & $\sigma$$=$$15$ & $\sigma$$=$$25$ & $\sigma$$=$$50$ \\
\midrule[0.15em]
IRCNN~\cite{zhang2017learning}   & 33.86 & 31.16 & 27.86 & 34.69 & 32.18 & 28.93 & 34.58 & 32.18 & 28.91 & 33.78 & 31.20 & 27.70\\
FFDNet~\cite{FFDNetPlus}  &33.87 & 31.21 & 27.96 & 34.63 & 32.13 & 28.98 & 34.66 & 32.35 & 29.18 & 33.83 & 31.40 & 28.05 \\
DnCNN~\cite{DnCNN}  &33.90 & 31.24 & 27.95 & 34.60 & 32.14 & 28.95 & 33.45 & 31.52 & 28.62 & 32.98 & 30.81 & 27.59 \\
DSNet~\cite{peng2019dilated}  & 33.91 & 31.28 & 28.05 & 34.63 & 32.16 & 29.05 & 34.67 & 32.40 & 29.28 & - & - & -\\
DRUNet~\cite{zhang2021DPIR}  & 34.30 & 31.69 & \underline{28.51} & 35.31 & 32.89 & 29.86 & 35.40 & 33.14 & 30.08 & 34.81 & 32.60 & 29.61 \\
Restormer~\cite{zamir2022restormer} & \underline{34.39} & \underline{31.78} & \textbf{28.59} & \underline{35.44} & \underline{33.02} & \underline{30.00} & \underline{35.55} & \underline{33.31} & \underline{30.29} & \underline{35.06} & \underline{32.91} & \underline{30.02}\\ 
\textbf{CSformer}$^*$ & \textbf{34.40} & \textbf{31.79} & \textbf{28.59} & \textbf{35.48} & \textbf{33.06} & \textbf{30.04} & \textbf{35.59} & \textbf{33.36} & \textbf{30.33} & \textbf{35.13} & \textbf{33.02} & \textbf{30.17}\\ 
\midrule[0.1em]
\midrule[0.1em]
RPCNN~\cite{xia2020rpcnn}  &- & 31.24 & 28.06 & - & 32.34 & 29.25 & - & 32.33 & 29.33 & - & 31.81 & 28.62\\
BRDNet~\cite{tian2020BRDnet}  &34.10 & 31.43 & 28.16 & 34.88 & 32.41 & 29.22 & 35.08 & 32.75 & 29.52 & 34.42 & 31.99 & 28.56 \\
RNAN~\cite{zhang2019residual}  &-&-&28.27&-&-&29.58&-&-&29.72&-&-&29.08\\
RDN~\cite{zhang2020rdn}  &-&-&28.31&-&-&29.66&-&-&-&-&-&29.38\\
IPT~\cite{chen2021IPT}  &-&-&28.39&-&-&29.64&-&-&29.98&-&-&29.71\\
SwinIR~\cite{liang2021swinir} & \textbf{34.42} & 31.78 & \underline{28.56} & 35.34 & 32.89 & 29.79 & 35.61 & 33.20 & 30.22 & 35.13 & 32.90 & 29.82 \\
Restormer~\cite{zamir2022restormer} & \underline{34.40} & \underline{31.79}& \textbf{28.60}& \underline{35.47} & \underline{33.04}& \underline{30.01}& \underline{35.61}& \underline{33.34}& \underline{30.30} & \underline{35.13}& \underline{32.96}& \underline{30.02}\\ 
\textbf{CSformer}$^*$ & \textbf{34.42} & \textbf{31.81}& \textbf{28.60}& \textbf{35.50} & \textbf{33.09}& \textbf{30.06}& \textbf{35.63}& \textbf{33.39}& \textbf{30.35} & \textbf{35.20}& \textbf{33.08}& \textbf{30.19}\\ 
\bottomrule[0.1em]
\end{tabular}}
}
\vspace*{-0.28em}
\end{table*}

\begin{table*}[t]
\begin{center}
\caption{\textbf{Real image denoising} results on SIDD~\cite{sidd} and DND~\cite{dnd} datasets.
}
\label{table:denoising_real}
\vspace{0.28em}
\setlength{\tabcolsep}{0.25em}
\scalebox{0.70}{
\begin{tabular}{c| c| c c c c c c c c c c c c c c c c}
\toprule[0.15em]
& \textbf{ Method} & RIDNet  & AINDNet  & VDN & SADNet &DANet+ & CycleISP & MIRNet & DeamNet & MPRNet & DAGL &  Uformer & MAXIM & Restormer & \textbf{CSformer} & \textbf{CSformer}$^*$ \\
\textbf{Dataset} & & \cite{RIDNet} & \cite{kim2020aindnet} & \cite{VDN} &  \cite{chang2020sadnet}	 & \cite{yue2020danet} &  \cite{zamir2020cycleisp} & \cite{zamir2020mirnet} & \cite{ren2021adaptivedeamnet}& \cite{Zamir_2021_CVPR_mprnet} & \cite{mou2021dynamicDAGL} & \cite{wang2021uformer} & \cite{tu2022maxim} & \cite{zamir2022restormer} & (Ours) & (Ours) \\
\midrule[0.15em]
\textbf{SIDD} & PSNR~$\textcolor{black}{\uparrow}$  &  38.71  &  39.08  & 39.28  & 39.46  & 39.47 & 39.52  & 39.72  & 39.47  & 39.71 & 38.94 & 39.89 & 39.96 & \underline{40.02} & 40.00 & \textbf{40.05}\\
~\cite{sidd}  & SSIM~$\textcolor{black}{\uparrow}$  &  0.951  &  0.954  & 0.956  & 0.957  & 0.957 & 0.957  & 0.959  & 0.957  & 0.958 & 0.953 & 0.960 & 0.960 & \underline{0.960} & \underline{0.960} & \textbf{0.961}\\
\midrule[0.1em]
\textbf{DND} & PSNR~$\textcolor{black}{\uparrow}$  &  39.26  &  39.37  & 39.38  & 39.59  & 39.58 & 39.56  & 39.88  & 39.63  & 39.80 & 39.77 & 39.98 & 39.84 & \textbf{40.03} & 40.00 & \underline{40.02} \\
~\cite{dnd}  & SSIM~$\textcolor{black}{\uparrow}$  &  0.953  &  0.951  & 0.952  & 0.952  & 0.955 & 0.956  & 0.956  & 0.953  & 0.954 & 0.956 & 0.955 & 0.954 & \textbf{0.956} & \textbf{0.956} & \textbf{0.956} \\
\bottomrule
\end{tabular}}
\end{center}
\vspace{-1.2em}
\end{table*}


\begin{figure*}[!t]
\begin{center}
\setlength{\tabcolsep}{0.15em}
\scalebox{0.97}{
\begin{tabular}[b]{c c c c c c c c}
\adjincludegraphics[trim={ 0 0 0 {.2\height} },clip,width=.122\textwidth,valign=t]{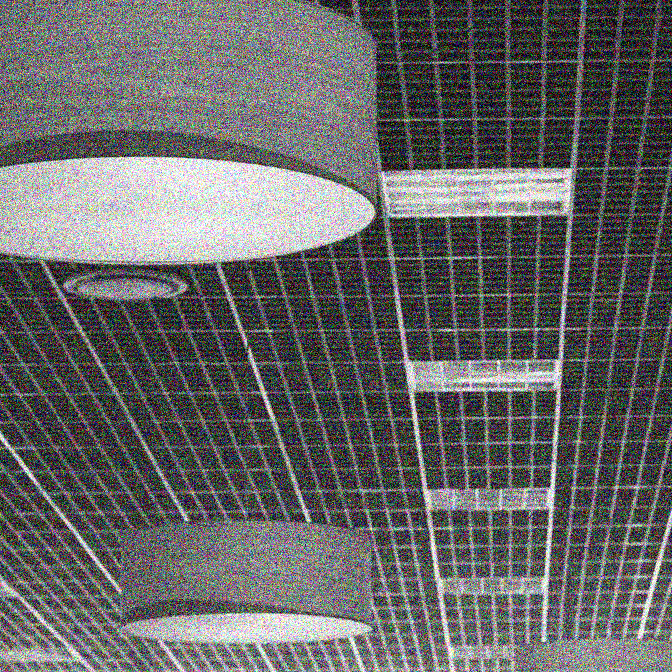} &  
\adjincludegraphics[trim={ 0 0 0 {.2\height} },clip,width=.122\textwidth,valign=t]{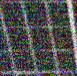} &  
\adjincludegraphics[trim={ 0 0 0 {.2\height} },clip,width=.122\textwidth,valign=t]{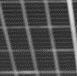} &  
\adjincludegraphics[trim={ 0 0 0 {.2\height} },clip,width=.122\textwidth,valign=t]{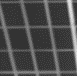} &   
\adjincludegraphics[trim={ 0 0 0 {.2\height} },clip,width=.122\textwidth,valign=t]{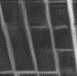} &   
\adjincludegraphics[trim={ 0 0 0 {.2\height} },clip,width=.122\textwidth,valign=t]{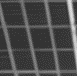} &  
\adjincludegraphics[trim={ 0 0 0 {.2\height} },clip,width=.122\textwidth,valign=t]{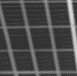} &  
\adjincludegraphics[trim={ 0 0 0 {.2\height} },clip,width=.122\textwidth,valign=t]{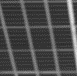} 
\\
\small~Noisy~image & \small~Noisy~patch & \small~Reference & \small~DRUNet~\cite{zhang2021DPIR} & \small~FFDNet~\cite{FFDNetPlus} & \small~SwinIR~\cite{liang2021swinir}  & \small~Restormer~\cite{zamir2022restormer} & \small~\textbf{CSformer}$^*$
\\
\adjincludegraphics[trim={ 0 {.1\height} 0 {.1\height} },clip,width=.122\textwidth,valign=t]{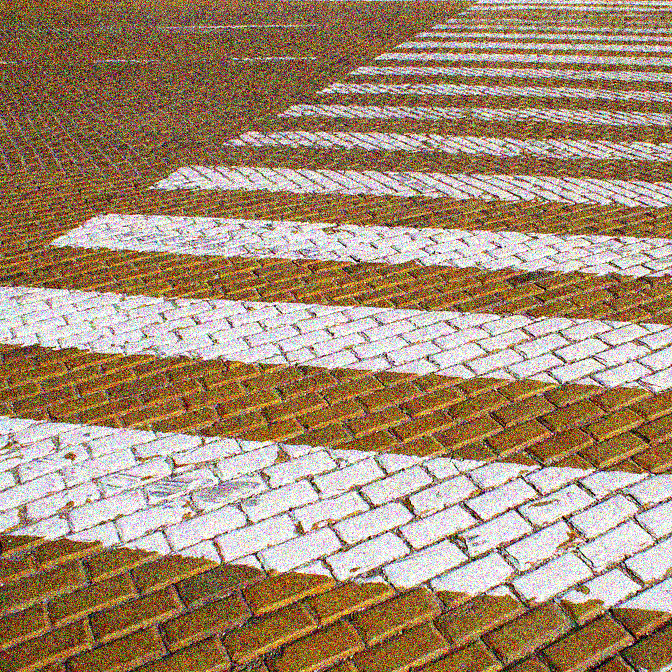} &   
\adjincludegraphics[trim={ 0 {.1\height} 0 {.1\height} },clip,width=.122\textwidth,valign=t]{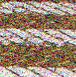} &   
\adjincludegraphics[trim={ 0 {.1\height} 0 {.1\height} },clip,width=.122\textwidth,valign=t]{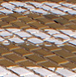} &  
\adjincludegraphics[trim={ 0 {.1\height} 0 {.1\height} },clip,width=.122\textwidth,valign=t]{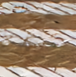} &   
\adjincludegraphics[trim={ 0 {.1\height} 0 {.1\height} },clip,width=.122\textwidth,valign=t]{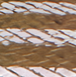} &   
\adjincludegraphics[trim={ 0 {.1\height} 0 {.1\height} },clip,width=.122\textwidth,valign=t]{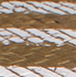} &  
\adjincludegraphics[trim={ 0 {.1\height} 0 {.1\height} },clip,width=.122\textwidth,valign=t]{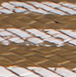} &  
\adjincludegraphics[trim={ 0 {.1\height} 0 {.1\height} },clip,width=.122\textwidth,valign=t]{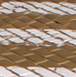} 
\\
\small~Noisy~image & \small~Noisy~patch & \small~Reference & \small~FFDNet~\cite{FFDNetPlus} & \small~IPT~\cite{chen2021IPT} & \small~SwinIR~\cite{liang2021swinir}  & \small~Restormer~\cite{zamir2022restormer} & \small~\textbf{CSformer}$^*$
\\
\includegraphics[width=.122\textwidth,valign=t]{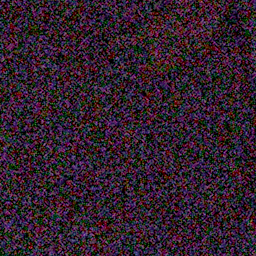} &   
\includegraphics[width=.122\textwidth,valign=t]{figures/images/denoising/real_input.png} &   
\includegraphics[width=.122\textwidth,valign=t]{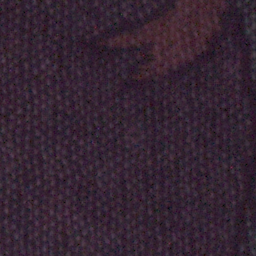} &  
\includegraphics[width=.122\textwidth,valign=t]{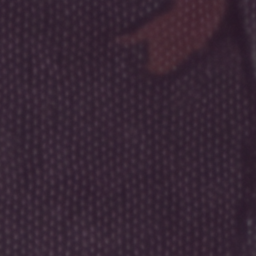} &   
\includegraphics[width=.122\textwidth,valign=t]{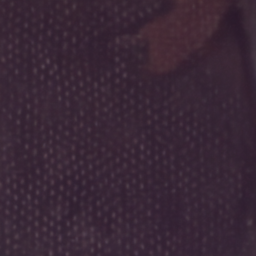} &   
\includegraphics[width=.122\textwidth,valign=t]{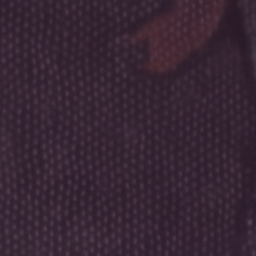} &  
\includegraphics[width=.122\textwidth,valign=t]{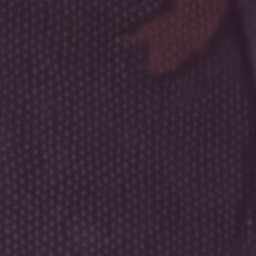} &  
\includegraphics[width=.122\textwidth,valign=t]{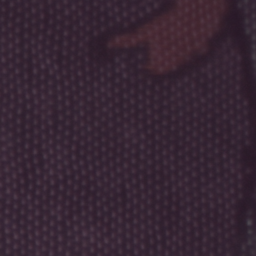} 
\\
\small~Noisy~image & \small~Noisy~patch & \small~Reference & \small~MIRNet~\cite{zamir2020mirnet} & \small~DeamNet~\cite{ren2021adaptivedeamnet} & \small~MPRNet~\cite{Zamir_2021_CVPR_mprnet}  & \small~Uformer~\cite{wang2021uformer} & \small~\textbf{CSformer}$^*$
\\
\end{tabular}}
\end{center}
\vspace{-0.75em}
\caption{\textbf{Image denoising} visualization comparisons. 
Top row: grayscale Gaussian noise removal on Urban100 \cite{huang2015single_urban100} ($\sigma=50$).
Middle row: color image Gaussian noise removal on Urban100 \cite{huang2015single_urban100} ($\sigma=50$).  
Bottom row: real image denoising on SIDD \cite{sidd}. 
}
\label{fig:denoising}
\vspace{-0.73em}
\end{figure*}

\vspace{-0.5em}
\section{Experiments}
\vspace{-0.5em}

We conduct extensive experiments to validate the effectiveness of the proposed model and the MAEIP pre-training framework.
We evaluate the proposed approach on a wide range of different tasks, including denoising, deblurring, and deraining.
More experiments on other tasks such as image enhancement \textit{etc.}, and more comprehensive results and visualizations can be found in the \textit{supplementary material}.
We \textbf{bold} the best results and \underline{underline} the second-best results in the tables.

\subsection{Experimental Setup}
\vspace{-0.5em}

\paragraph{Architectural configuration.} We introduce three CSformer variants in our experiments including CSformer-T (Tiny), CSformer-T2, and CSformer by setting different base feature channels $C$ and the number of CSformer blocks in each encoder and decoder stage.
The design of the encoder and decoder in CSformer adopts a symmetrical structure.
The detailed architecture for each variant is listed in the supplementary material.

\paragraph{Training details.}
Our proposed CSformer is end-to-end trainable without pre-training \cite{chen2021IPT} or progressive training \cite{mspfn2020,zamir2022restormer}.
However, with the help of self-supervised MAEIP pre-training, the performance can be further improved.
In the following experiments, we use $^*$ to indicate a CSformer that has been pre-trained with MAEIP before finetuning, \textit{i.e.,} CSformer$^*$.
We perform self-supervised pre-training on the ImageNet-1K (IN1K) \cite{deng2009imagenet} training set with $192 \times 192$ random-cropped patches.
Due to the limitation of computing resources, we only pre-train MAEIP for 100 epochs on IN1K.
For two-stage pre-training, we first pre-train the encoder for 50 epochs and then pre-train the whole network for 50 epochs.
After MAEIP pre-training, we finetune separate models for different image processing tasks.
Generally, the CSformer is trained on $192 \times 192$ random-cropped patches with random data augmentation methods including horizontal and vertical flips, $90^{\circ}$ rotations, and MixUp \cite{zhang2017mixup}.
For the tasks that highly depend on long-range dependencies and usually have high-resolution inputs, \textit{i.e.,} motion deblurring and defocus deblurring, the CSformer is finetuned on $256 \times 256$ random-cropped patches.
We generally train CSformer using the AdamW optimizer for 600K iterations with an initial learning rate of $2e^{-4}$, and gradually reduce the learning rate to $1e^{-6}$ with the cosine annealing \cite{loshchilovsgdr}.
Without specific instructions, we test the CSformer on full-resolution images rather than on small patches.

\begin{table}[!t]
\begin{center}
\vspace{0.2em}
\caption{\textbf{Motion deblurring} results. Our CSformer$^*$ is only trained on the GoPro training set~\cite{gopro2017}, and directly evaluated on the GoPro test set~\cite{gopro2017}, HIDE~\cite{shen2019human}, and RealBlur-R\&J~\cite{rim_2020_realblur}}
\label{table:deblurring_motion}
\vspace{-0.4em}
\setlength{\tabcolsep}{0.19em}
\scalebox{0.69}{
\begin{tabular}{l c | c | c | c }
\toprule[0.15em]
 & \textbf{GoPro}~\cite{gopro2017} & \textbf{HIDE}~\cite{shen2019human} & \textbf{RealBlur-R}~\cite{rim_2020_realblur} & \textbf{\textbf{RealBlur-J}}~\cite{rim_2020_realblur} \\
 \textbf{Method} & PSNR~\colorbox{color4}{SSIM} & PSNR~\colorbox{color4}{SSIM} & PSNR~\colorbox{color4}{SSIM} & PSNR~\colorbox{color4}{SSIM}\\
\midrule[0.15em]
Xu \etal \cite{xu2013unnatural}     & 21.00 \colorbox{color4}{0.741} & -  &   34.46 \colorbox{color4}{0.937} &  27.14 \colorbox{color4}{0.830} \\
DeblurGAN \cite{deblurgan}          & 28.70 \colorbox{color4}{0.858} & 24.51 \colorbox{color4}{0.871} & 33.79 \colorbox{color4}{0.903}  &  27.97 \colorbox{color4}{0.834} \\
Nah \etal \cite{gopro2017}          & 29.08 \colorbox{color4}{0.914} & 25.73 \colorbox{color4}{0.874}  &  32.51 \colorbox{color4}{0.841}  &  27.87 \colorbox{color4}{0.827} \\
Zhang \etal \cite{zhang2018dynamic} & 29.19 \colorbox{color4}{0.931} & - &  35.48 \colorbox{color4}{0.947}  &  27.80 \colorbox{color4}{0.847}\\
\small{DeblurGAN-v2 \cite{deblurganv2}}    & 29.55 \colorbox{color4}{0.934} & 26.61 \colorbox{color4}{0.875} & 35.26 \colorbox{color4}{0.944}  &  28.70 \colorbox{color4}{0.866} \\
SRN~\cite{tao2018scale}             & 30.26 \colorbox{color4}{0.934} & 28.36 \colorbox{color4}{0.915} & 35.66 \colorbox{color4}{0.947} &  28.56  \colorbox{color4}{0.867} \\
Shen \etal \cite{shen2019human}     & -                         & 28.89 \colorbox{color4}{0.930} & -        & -  \\
Gao \etal \cite{gao2019dynamic}     & 30.90 \colorbox{color4}{0.935} &  29.11 \colorbox{color4}{0.913}  & -      & - \\
DBGAN \cite{zhang2020dbgan}         & 31.10 \colorbox{color4}{0.942} & 28.94 \colorbox{color4}{0.915}  & 33.78 \colorbox{color4}{0.909}     & 24.93 \colorbox{color4}{0.745} \\
MT-RNN \cite{mtrnn2020}             & 31.15 \colorbox{color4}{0.945} & 29.15 \colorbox{color4}{0.918}   & 35.79 \colorbox{color4}{0.951}     & 28.44 \colorbox{color4}{0.862}\\
DMPHN \cite{dmphn2019}              & 31.20 \colorbox{color4}{0.940}  & 29.09 \colorbox{color4}{0.924} &  35.70 \colorbox{color4}{0.948} & 28.42 \colorbox{color4}{0.860} \\
Suin \etal \cite{Maitreya2020}      & 31.85 \colorbox{color4}{0.948} & 29.98 \colorbox{color4}{0.930} & -       & - \\
SPAIR~\cite{purohit2021spatially_spair} & 32.06 \colorbox{color4}{0.953} & 30.29 \colorbox{color4}{0.931} & - & \underline{28.81} \colorbox{color4}{\underline{0.875}}\\
MIMO-UNet+~\cite{cho2021rethinking_mimo} & 32.45 \colorbox{color4}{0.957} & 29.99 \colorbox{color4}{0.930} & 35.54 \colorbox{color4}{0.947} & 27.63 \colorbox{color4}{0.837}\\
IPT~\cite{chen2021IPT} & \hspace{-2em} 32.52 \colorbox{color4}{-} & - & - & -\\
MPRNet~\cite{Zamir_2021_CVPR_mprnet} & 32.66 \colorbox{color4}{0.959} &	30.96 \colorbox{color4}{0.939} & 35.99   \colorbox{color4}{0.952} & 28.70 \colorbox{color4}{0.873} \\
Restormer \cite{zamir2022restormer} & \underline{32.92} \colorbox{color4}{\underline{0.961}} & \underline{31.22} \colorbox{color4}{\underline{0.942}} & \underline{36.19} \colorbox{color4}{\underline{0.957}} & \textbf{28.96} \colorbox{color4}{\textbf{0.879}}\\
\bottomrule[0.1em]
\textbf{CSformer}$^*$ & \textbf{33.36} \colorbox{color4}{\textbf{0.965}} & \textbf{31.39} \colorbox{color4}{\textbf{0.945}} & \textbf{36.29} \colorbox{color4}{\textbf{0.958}} & \underline{28.81} \colorbox{color4}{0.873}\\
\bottomrule[0.1em]
\end{tabular}}
\end{center}\vspace{-2.2em}
\end{table}

\begin{figure*}[!t]
\begin{center}
\setlength{\tabcolsep}{0.15em}
\scalebox{0.97}{
\begin{tabular}[b]{c c c c c c c c}
\includegraphics[trim={ 0 0 315 0
 },clip,width=.122\textwidth,valign=t]{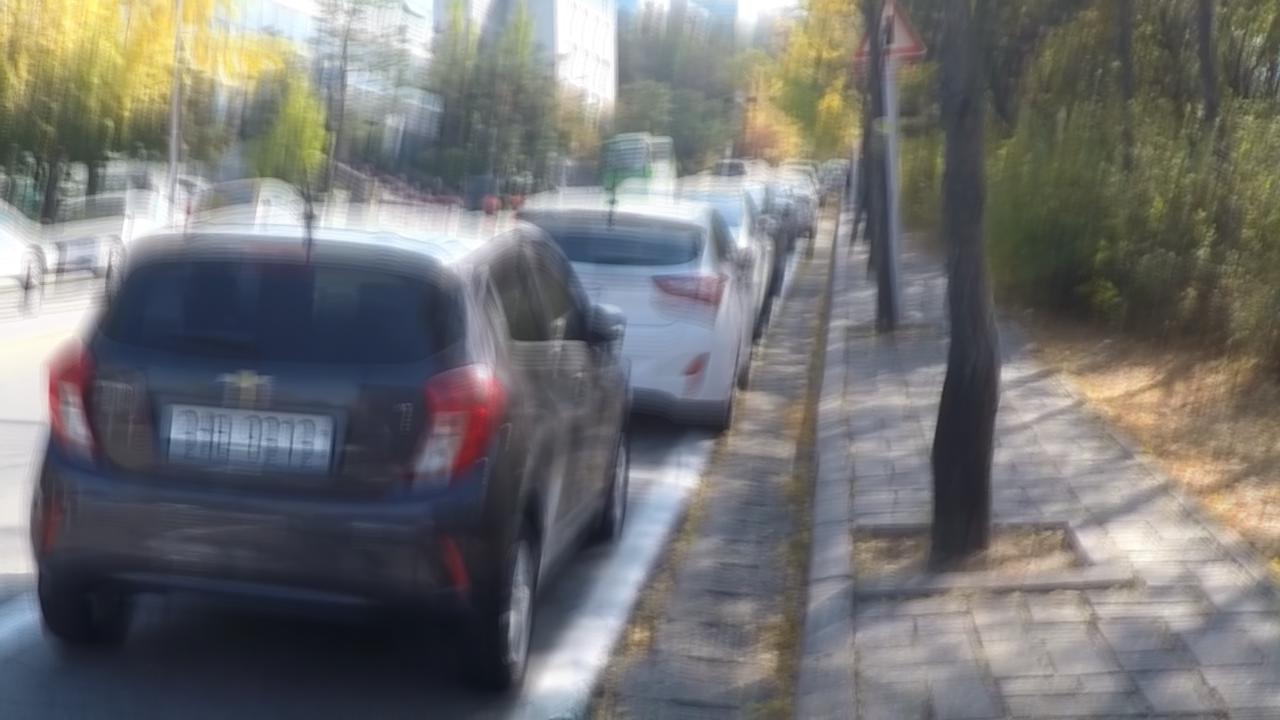} &   
\includegraphics[width=.122\textwidth,valign=t]{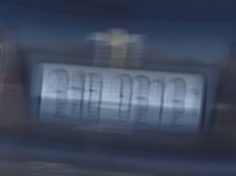} &   
\includegraphics[width=.122\textwidth,valign=t]{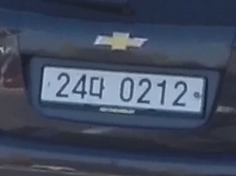} &  
\includegraphics[width=.122\textwidth,valign=t]{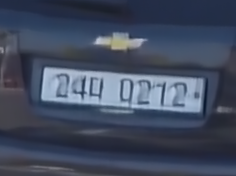} &   
\includegraphics[width=.122\textwidth,valign=t]{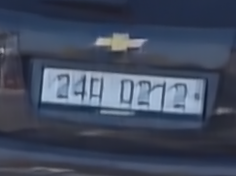} &   
\includegraphics[width=.122\textwidth,valign=t]{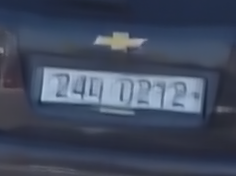} &  
\includegraphics[width=.122\textwidth,valign=t]{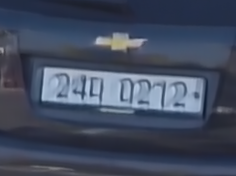} &  
\includegraphics[width=.122\textwidth,valign=t]{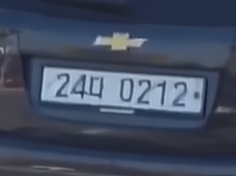} 
\vspace{0.3em}
\\
\includegraphics[trim={ 250 0 0 0
 },clip,width=.122\textwidth,valign=t]{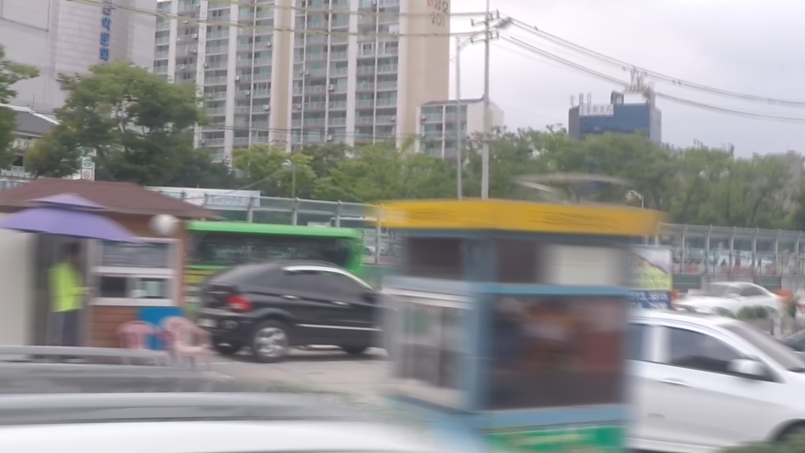} &   
\includegraphics[width=.122\textwidth,valign=t]{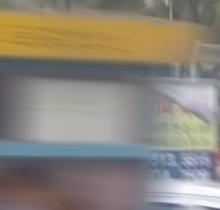} &   
\includegraphics[width=.122\textwidth,valign=t]{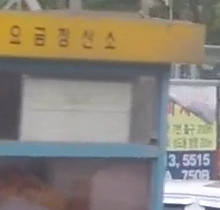} &  
\includegraphics[width=.122\textwidth,valign=t]{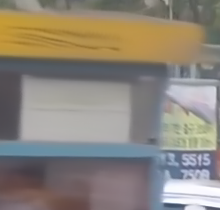} &   
\includegraphics[width=.122\textwidth,valign=t]{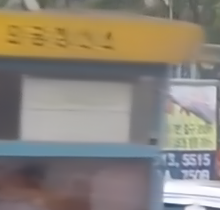} &   
\includegraphics[width=.122\textwidth,valign=t]{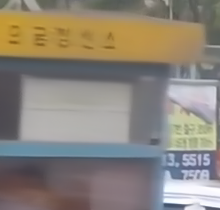} &  
\includegraphics[width=.122\textwidth,valign=t]{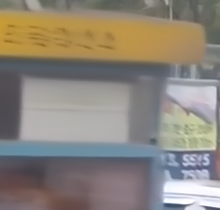} &  
\includegraphics[width=.122\textwidth,valign=t]{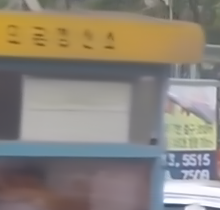} 
\\
\small~Blurry~image & \small~Blurry~patch & \small~Reference & \small~MPRNet~\cite{Zamir_2021_CVPR_mprnet} & \small~MIMO+~\cite{cho2021rethinking_mimo} & \small~MAXIM~\cite{tu2022maxim}  & \small~Restormer~\cite{zamir2022restormer} & \small~\textbf{CSformer}$^*$
\\
\end{tabular}}
\end{center}
\vspace{-0.8em}
\caption{\textbf{Motion deblurring} results on GoPro \cite{gopro2017} test set. Our CSformer$^*$ effectively removes blur and generates more faithful results. 
}
\label{fig:deblurring_motion_gopro}
\vspace{-0.8em}
\end{figure*}

\vspace{-0.3em}
\subsection{Benchmark Results}
\vspace{-0.3em}

\paragraph{Gaussian denoising.} We first conduct denoising experiments on synthetic datasets with additive white Gaussian noise.
Following previous conventions \cite{DnCNN,liang2021swinir,zamir2022restormer}, we train the gaussian denoising model on four datasets including DIV2K~\cite{agustsson2017ntire}, Flickr2K, BSD500~\cite{arbelaez2010contour}, WaterlooED~\cite{ma2016waterloo}, and test on five datasets including Set12 \cite{DnCNN}, BSD68 \cite{martin2001database_bsd}, Urban100 \cite{huang2015single_urban100}, Kodak24 \cite{kodak}, and McMaster \cite{zhang2011color_mcmaster}.
We consider three noise levels including 15, 25, and 50.
We conduct two types of gaussian denoising experiments, including learning a single denoising model to deal with various noise levels (blind denoising) and learning separate models for different noise levels respectively (non-blind denoising).
Table \ref{table:graydenoising} and Table \ref{table:colordenoising} show PSNR results of different models for grayscale image denoising and color image denoising, respectively.
Our CSformer$^*$ achieves state-of-the-art performance compared to other models.
Specifically, on the grayscale Urban100 \cite{huang2015single_urban100} dataset with a noise level of 50, CSformer$^*$ yields 0.23 dB gain over Restormer \cite{zamir2022restormer} and 0.54 dB gain over SwinIR \cite{liang2021swinir}, respectively, for the non-blind denoising task.
The top row and middle row in Figure \ref{fig:denoising} show visualization results of different denoising models.
Our CSformer$^*$ produces images that are more conformed to human visual preference.

\paragraph{Real denoising.} We further conduct denoising experiments on real-world datasets, \textit{i.e.,} SIDD \cite{sidd} and DND \cite{dnd}.
Following the previous conventions \cite{wang2021uformer,tu2022maxim,zamir2022restormer}, we only train CSformer on the SIDD \cite{sidd} training set and directly test on the SIDD \cite{sidd} and DND \cite{dnd} test sets.
Quantitative results in Table \ref{table:denoising_real} demonstrate that our CSformer$^*$ achieves state-of-the-art performance on the SIDD dataset and obtains competitive performance on the DND dataset.
Moreover, without MAEIP pre-training, our CSformer still outperforms most of the SOTA models, which also manifests the superiority of the CSformer.
The bottom row in Figure \ref{fig:denoising} also qualitatively demonstrates the effectiveness of our CSformer$^*$ on real denoising.

\begin{table*}[!t]
\begin{center}
\caption{\textbf{Defocus deblurring} comparisons of the state-of-the-art models on the DPDD test set~\cite{abdullah2020dpdd}, which contains 37 indoor scenes and 39 outdoor scenes. \textbf{S:} single-image defocus deblurring. \textbf{D:} dual-pixel defocus deblurring.
}
\label{table:deblurring_defocus}
\vspace{-0.4em}
\setlength{\tabcolsep}{1em}
\scalebox{0.7}{
\begin{tabular}{l | c c c c | c c c c | c c c c }
\toprule[0.15em]
   & \multicolumn{4}{c|}{\textbf{Indoor Scenes}} & \multicolumn{4}{c|}{\textbf{Outdoor Scenes}} & \multicolumn{4}{c}{\textbf{Combined}} \\
\cline{2-13}
   \textbf{Method} & PSNR~$\textcolor{black}{\uparrow}$ & SSIM~$\textcolor{black}{\uparrow}$& MAE~$\textcolor{black}{\downarrow}$ & LPIPS~$\textcolor{black}{\downarrow}$  & PSNR~$\textcolor{black}{\uparrow}$ & SSIM~$\textcolor{black}{\uparrow}$& MAE~$\textcolor{black}{\downarrow}$ & LPIPS~$\textcolor{black}{\downarrow}$  & PSNR~$\textcolor{black}{\uparrow}$ & SSIM~$\textcolor{black}{\uparrow}$& MAE~$\textcolor{black}{\downarrow}$ & LPIPS~$\textcolor{black}{\downarrow}$   \\
\midrule[0.15em]
EBDB$_S$~\cite{karaali2017edge_EBDB} & 25.77 & 0.772 & 0.040 & 0.297 & 21.25 & 0.599 & 0.058 & 0.373 & 23.45 & 0.683 & 0.049 & 0.336 \\
DMENet$_S$~\cite{lee2019deep_dmenet}  & 25.50 & 0.788 & 0.038 & 0.298 & 21.43 & 0.644 & 0.063 & 0.397 & 23.41 & 0.714 & 0.051 & 0.349 \\
JNB$_S$~\cite{shi2015just_jnb} & 26.73 & 0.828 & 0.031 & 0.273 & 21.10 & 0.608 & 0.064 & 0.355 & 23.84 & 0.715 & 0.048 & 0.315 \\
DPDNet$_S$~\cite{abdullah2020dpdd} &26.54 & 0.816 & 0.031 & 0.239 & 22.25 & 0.682 & 0.056 & 0.313 & 24.34 & 0.747 & 0.044 & 0.277\\
KPAC$_S$~\cite{son2021single_kpac} & 27.97 & 0.852 & 0.026 & 0.182 & 22.62 & 0.701 & 0.053 & 0.269 & 25.22 & 0.774 & 0.040 & 0.227 \\
IFAN$_S$~\cite{Lee_2021_CVPRifan} & 28.11  & 0.861  & 0.026 & 0.179  & 22.76  & 0.720 & 0.052  & 0.254  & 25.37 & 0.789 & 0.039 & 0.217\\
Restormer$_S$~\cite{zamir2022restormer}& \underline{28.87}  & \underline{0.882}  & \underline{0.025} & \underline{0.145} & \underline{23.24}  & \underline{0.743}  & \underline{0.050} & \underline{0.209}  & \underline{25.98}  & \underline{0.811}  & \underline{0.038}  & \underline{0.178}   \\
\textbf{CSformer}$^*_S$ & \textbf{29.01}  & \textbf{0.883}  & \textbf{0.023} & \textbf{0.139} & \textbf{23.63}  & \textbf{0.759}  & \textbf{0.047} & \textbf{0.191}  & \textbf{26.25}  & \textbf{0.819}  & \textbf{0.036}  & \textbf{0.166}   \\
\midrule[0.1em]
\midrule[0.1em]
DPDNet$_D$~\cite{abdullah2020dpdd} & 27.48 & 0.849 & 0.029 & 0.189 & 22.90 & 0.726 & 0.052 & 0.255 & 25.13 & 0.786 & 0.041 & 0.223 \\
RDPD$_D$~\cite{abdullah2021rdpd} & 28.10 & 0.843 & 0.027 & 0.210 & 22.82 & 0.704 & 0.053 & 0.298 & 25.39 & 0.772 & 0.040 & 0.255 \\

Uformer$_D$~\cite{wang2021uformer} & 28.23 & 0.860 & 0.026 & 0.199 & 23.10 & 0.728 & 0.051 & 0.285 & 25.65 & 0.795 & 0.039 & 0.243 \\

IFAN$_D$~\cite{Lee_2021_CVPRifan} & 28.66 & 0.868 & \underline{0.025} & 0.172 & 23.46 & 0.743 & 0.049 & 0.240 & 25.99 & 0.804 & 0.037 & \underline{0.207} \\

Restormer$_D$~\cite{zamir2022restormer}& \underline{29.48}  & \underline{0.895}  & \textbf{0.023} & \underline{0.134} & \underline{23.97}  & \underline{0.773}  & \underline{0.047} & \textbf{0.175}  & \underline{26.66}  & \underline{0.833}  & \underline{0.035}  & \textbf{0.155} \\

\textbf{CSformer}$^*_D$& \textbf{29.54}  & \textbf{0.896}  & \textbf{0.023} & \textbf{0.130} & \textbf{24.38}  & \textbf{0.788}  & \textbf{0.045} & \underline{0.179}  & \textbf{26.89}  & \textbf{0.841}  & \textbf{0.034}  & \textbf{0.155} \\
\bottomrule[0.1em]
\end{tabular}}
\end{center}
\vspace{-1em}
\end{table*}

\begin{figure*}[!t]
\begin{center}
\setlength{\tabcolsep}{0.15em}
\scalebox{0.97}{
\begin{tabular}[b]{c c c c c c c c}
\includegraphics[width=.122\textwidth,valign=t]{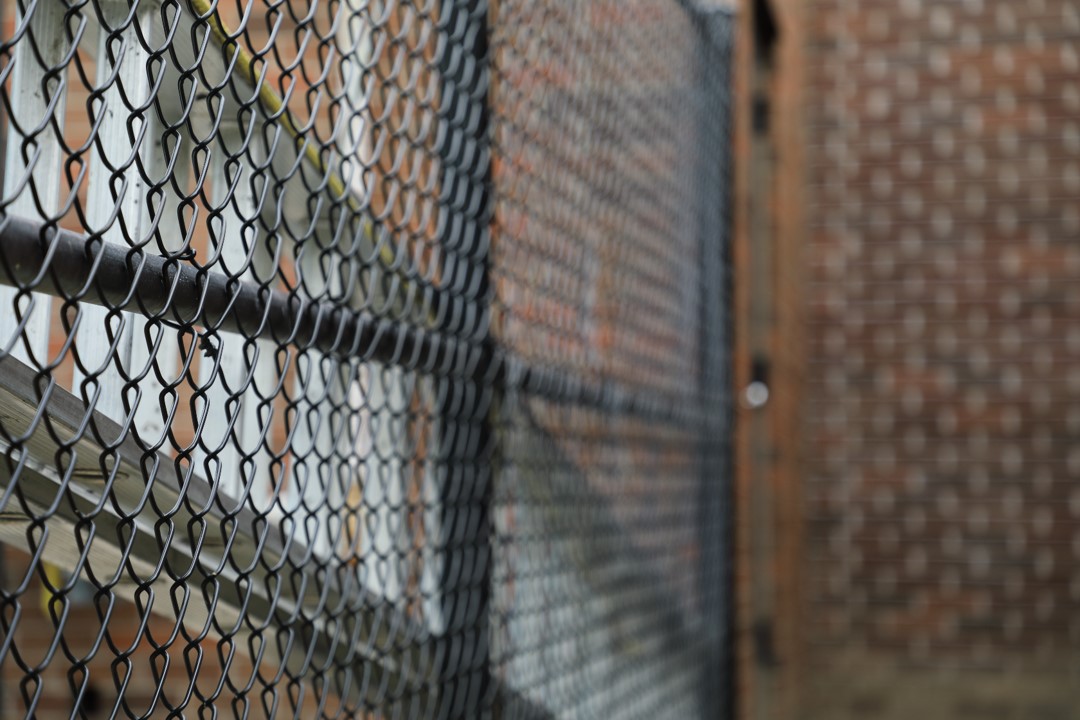} &   
\includegraphics[width=.122\textwidth,valign=t]{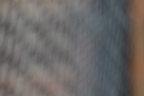} &   
\includegraphics[width=.122\textwidth,valign=t]{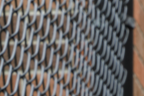} &  
\includegraphics[width=.122\textwidth,valign=t]{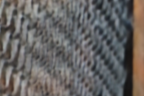} &   
\includegraphics[width=.122\textwidth,valign=t]{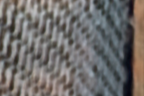} &   
\includegraphics[width=.122\textwidth,valign=t]{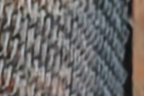} &  
\includegraphics[width=.122\textwidth,valign=t]{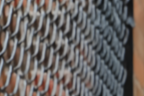} &  
\includegraphics[width=.122\textwidth,valign=t]{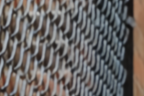} 
\vspace{0.3em}
\\
\includegraphics[width=.122\textwidth,valign=t]{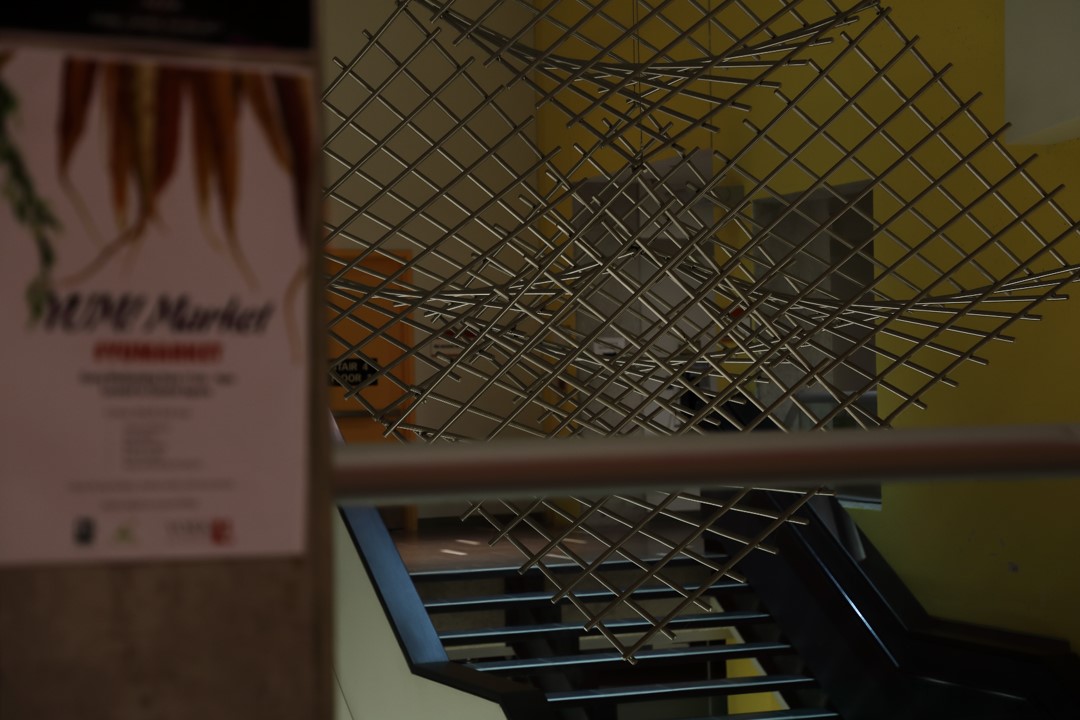} &   
\includegraphics[width=.122\textwidth,valign=t]{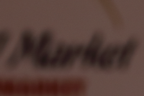} &   
\includegraphics[width=.122\textwidth,valign=t]{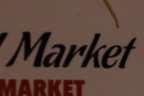} &  
\includegraphics[width=.122\textwidth,valign=t]{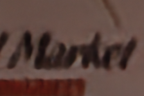} &   
\includegraphics[width=.122\textwidth,valign=t]{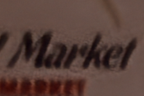} &   
\includegraphics[width=.122\textwidth,valign=t]{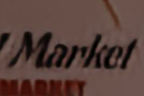} &  
\includegraphics[width=.122\textwidth,valign=t]{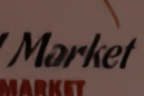} &  
\includegraphics[width=.122\textwidth,valign=t]{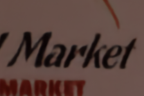} 
\\
\small~Blurry~image & \small~Blurry~patch & \small~Reference & \small~DPDNet~\cite{abdullah2020dpdd} & \small~RDPD~\cite{abdullah2021rdpd} & \small~IFAN~\cite{Lee_2021_CVPRifan}  & \small~Restormer~\cite{zamir2022restormer} & \small~\textbf{CSformer}$^*$
\\
\end{tabular}}
\end{center}
\vspace*{-0.8em}
\caption{\textbf{Defocus deblurring} results on the DPDD \cite{abdullah2020dpdd} test set. Our CSformer$^*$ effectively removes blur and generates sharper results. 
}
\label{fig:deblurring_defocus}
\vspace{-0.3em}
\end{figure*}

\paragraph{Motion deblurring.} Following previous experimental settings on motion deblurring \cite{Zamir_2021_CVPR_mprnet,wang2021uformer,zamir2022restormer}, we only train CSformer$^*$ on the GoPro \cite{gopro2017} training set and directly test the performance on four test sets, including two synthetic datasets (GoPro test set \cite{gopro2017}, HIDE \cite{shen2019human}) and two real-world datasets (RealBlur-R \cite{rim_2020_realblur}, RealBlur-J \cite{rim_2020_realblur}).
The CSformer$^*$ is tested on full-resolution images rather than on patches \cite{chen2022nafnet}.
Table \ref{table:deblurring_motion} shows the quantitative comparisons of different models on the four test sets.
Our CSformer$^*$ achieves a performance boost of 0.44 dB over the Restormer \cite{zamir2022restormer} on the GoPro test set.
Moreover, our CSformer$^*$ also generalizes well on the other three test sets.
Figure \ref{fig:deblurring_motion_gopro} demonstrates the visualization results of different deblurring models.
It can be observed that our CSformer$^*$ generates more visually-faithful results compared to other methods.

\paragraph{Defocus deblurring.} We conduct the defocus deblurring experiments on the DPDD dataset \cite{abdullah2020dpdd}.
We consider two cases including the single-image defocus deblurring and the dual-pixel defocus deblurring.
For dual-pixel defocus deblurring, we directly concatenate the left image and the right image in the channel dimension, and then input it to the CSformer$^*$.
Table \ref{table:deblurring_defocus} shows the quantitative comparisons of state-of-the-art models on the DPDD test set.
Our CSformer$^*$ outperforms previous models on almost all evaluation criteria for both tasks and both scenes.
Figure \ref{fig:deblurring_defocus} illustrates that our method is more effective in removing defocus blur compared to other approaches.

\begin{table*}
\begin{center}
\caption{Quantitative comparisons of the state-of-the-art models for \textbf{image deraining}.}
\label{table:deraining}
\vspace{0.4em}
\setlength{\tabcolsep}{1.03em}
\scalebox{0.7}{
\begin{tabular}{l | c c | c c | c c | c c | c c || c c}
\toprule[0.15em]
  & \multicolumn{2}{c|}{\textbf{Rain100L}~\cite{yang2017deep}}&\multicolumn{2}{c|}{\textbf{Rain100H}~\cite{yang2017deep}}&\multicolumn{2}{c|}{\textbf{Test100}~\cite{zhang2019image}}&\multicolumn{2}{c|}{\textbf{Test1200}~\cite{zhang2018density}}&\multicolumn{2}{c||}{\textbf{Test2800}~\cite{fu2017removing}}&\multicolumn{2}{c}{\textbf{Average}}\\
\textbf{Method} & PSNR$\uparrow$ & SSIM$\uparrow$ & PSNR$\uparrow$ & SSIM$\uparrow$  & PSNR$\uparrow$ & SSIM$\uparrow$ & PSNR$\uparrow$ & SSIM$\uparrow$  & PSNR$\uparrow$ & SSIM$\uparrow$  & PSNR$\uparrow$ & SSIM$\uparrow$ \\
\toprule
DerainNet~\cite{fu2017clearing} & 27.03 & 0.884 & 14.92 & 0.592 & 22.77 & 0.810 & 23.38 & 0.835 & 24.31 & 0.861 & 22.48 & 0.796  \\
SEMI~\cite{wei2019semi} & 25.03 & 0.842 & 16.56 & 0.486 & 22.35 & 0.788 & 26.05 & 0.822 & 24.43 & 0.782 & 22.88 & 0.744 \\
DIDMDN~\cite{zhang2018density} & 25.23 & 0.741 & 17.35 & 0.524 & 22.56 & 0.818 & 29.65 & 0.901 & 28.13 & 0.867 & 24.58 & 0.770 \\
UMRL~\cite{yasarla2019uncertainty} & 29.18 & 0.923 & 26.01 & 0.832 & 24.41 & 0.829 & 30.55 & 0.910 & 29.97 & 0.905 &   28.02 &  0.880\\ 
RESCAN~\cite{li2018recurrent} & 29.80 & 0.881 & 26.36 & 0.786 & 25.00 & 0.835 & 30.51 & 0.882 & 31.29 & 0.904 & 28.59 & 0.857 \\
PreNet~\cite{ren2019progressive} & 32.44 & 0.950 & 26.77 & 0.858 & 24.81 & 0.851 & 31.36 & 0.911 & 31.75 & 0.916 & 29.42 & 0.897 \\
MSPFN~\cite{mspfn2020} & 32.40 & 0.933 & 28.66 & 0.860 & 27.50 & 0.876 & 32.39 & 0.916 & 32.82 & 0.930 & 30.75 & 0.903 \\
MPRNet~\cite{Zamir_2021_CVPR_mprnet} & 36.40 & 0.965 & 30.41 & 0.890 & 30.27 & 0.897 & 32.91 & 0.916 & 33.64 & 0.938 & 32.73 & 0.921 \\
HINet~\cite{chen2021hinet} & 37.20 & 0.969 & 30.63 & 0.893 & 30.26 & 0.905 & \underline{33.01} & 0.918 & \underline{33.87} & 0.940 & 33.00 & 0.925 \\
MAXIM-2S~\cite{tu2022maxim} & \underline{38.06} &	\underline{0.977} & \underline{30.81} & \underline{0.903} & \underline{31.17} & \underline{0.922}	& 32.37 & \underline{0.922}	& 33.80 & \underline{0.943} & \underline{33.24} & \underline{0.933} \\
\midrule
\textbf{CSformer}$^*$ & \textbf{39.49} & \textbf{0.978} & \textbf{32.00} & \textbf{0.912} & \textbf{32.02} & \textbf{0.925} & \textbf{33.19} & \textbf{0.925} & \textbf{33.81} & \textbf{0.943} & \textbf{34.10} & \textbf{0.937} \\
\bottomrule[0.1em]
\end{tabular}}
\end{center}
\vspace{-0.8em}
\end{table*}


\begin{figure*}[!t]
\begin{center}
\setlength{\tabcolsep}{0.14em}
\scalebox{0.97}{
\begin{tabular}[b]{c c c c c c c c}
\includegraphics[width=.122\textwidth,valign=t]{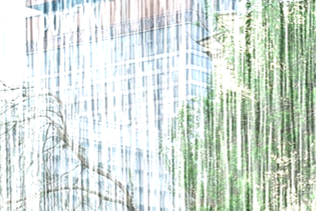} &   
\includegraphics[width=.122\textwidth,valign=t]{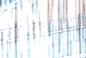} &   
\includegraphics[width=.122\textwidth,valign=t]{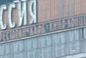} &  
\includegraphics[width=.122\textwidth,valign=t]{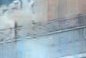} & 
\includegraphics[width=.122\textwidth,valign=t]{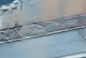} &  
\includegraphics[width=.122\textwidth,valign=t]{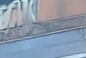} &  
\includegraphics[width=.122\textwidth,valign=t]{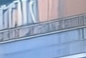} &   
\includegraphics[width=.122\textwidth,valign=t]{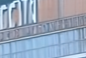}
\vspace{0.3em}
\\
\includegraphics[width=.122\textwidth,valign=t]{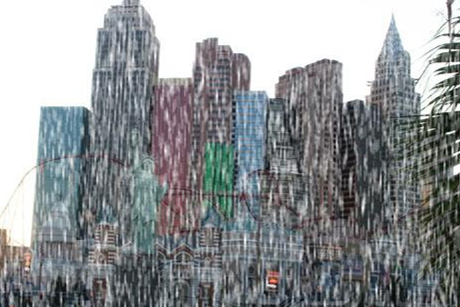} &   
\includegraphics[width=.122\textwidth,valign=t]{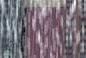} &   
\includegraphics[width=.122\textwidth,valign=t]{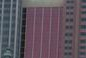} &   
\includegraphics[width=.122\textwidth,valign=t]{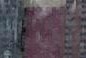} & 
\includegraphics[width=.122\textwidth,valign=t]{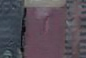} &  
\includegraphics[width=.122\textwidth,valign=t]{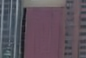} &  
\includegraphics[width=.122\textwidth,valign=t]{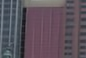} &   
\includegraphics[width=.122\textwidth,valign=t]{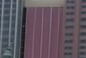}
\\
\small~Rainy~image & \small~Rainy~patch & \small~Reference & \small~RESCAN~\cite{li2018recurrent} & \small~PreNet~\cite{ren2019progressive} & \small~MPRNet~\cite{Zamir_2021_CVPR_mprnet}  & \small~MAXIM~\cite{tu2022maxim} & \small~\textbf{CSformer}$^*$
\\
\end{tabular}}
\end{center}
\vspace{-0.9em}
\caption{\textbf{Image deraining} visualization results. Our CSformer$^*$ produces more visually pleasant rain-free results. 
}
\label{fig:deraining}
\vspace{-0.4em}
\end{figure*}

\paragraph{Deraining.} Following previous works \cite{mspfn2020,Zamir_2021_CVPR_mprnet,tu2022maxim}, for single image deraining experiments, we train CSformer$^*$ on a comprehensive training set \cite{mspfn2020}, and test the performance on five test sets including Rain100L \cite{yang2017deep}, Rain100H \cite{yang2017deep}, Test100 \cite{zhang2019image}, Test1200 \cite{zhang2018density}, and Test 2800 \cite{fu2017removing}.
Table \ref{table:deraining} shows quantitative comparisons with several previous methods. It can be observed that our CSformer$^*$ achieves state-of-the-art performances on all test sets.
Figure \ref{fig:deraining} also gives some examples of the visualization results of different deraining methods.
Our method produces rain-free images with structural fidelity and without introducing noticeable artifacts.

\begin{table}[!t]
\centering
\scriptsize
\setlength{\tabcolsep}{0.18em}
\renewcommand{\arraystretch}{1.1}
\caption{Ablation studies of individual components in CSformer and the MAEIP pre-training method.}
\vspace{0.8em}
\begin{tabular}[t]{@{}c@{}}

\scriptsize
\setlength{\tabcolsep}{0.11em}
\begin{tabular}{x{12}x{12}x{12}x{12}|x{25}x{20}}
 \toprule
 \scriptsize{GC} &  \scriptsize{Sym} &  \scriptsize{GA} &  \scriptsize{CA} &  \scriptsize{GMACs} &  \scriptsize{PSNR}   \\
\midrule
       &        &        &        &    12.00    & 39.66 \\ \arrayrulecolor{lightgray}\hline
\cmark &        &        &        &    11.69    & 39.68 \\ \arrayrulecolor{lightgray}\hline
\cmark & \cmark &        &        &    12.66    & 39.72 \\ \arrayrulecolor{lightgray}\hline
\cmark & \cmark & \cmark &        &    12.66    & 39.74 \\ \arrayrulecolor{lightgray}\hline 
\cmark & \cmark & \cmark & \cmark &    13.50    & 39.77 \\ 
\arrayrulecolor{black}\bottomrule
\specialrule{0em}{0.3em}{0em}
\multicolumn{6}{c}{{A. Individual components.}}\\
\end{tabular}

\hfill

\scriptsize
\setlength{\tabcolsep}{0.19em}
\def\arraystretch{1.13}
\begin{tabular}{l|x{28}|x{26}|x{25}}
 \toprule
Variant & Rain100L & Test100 & Avg. \\
\midrule
w/o p. & 37.03  & 30.53 & 33.78 \\ 
r. encoder & 37.11 & 30.90 & 34.00 \\ 
r. decoder & 37.35 & 30.99 & 34.17 \\ 
MAEIP & 37.48 & 31.07 & 34.28 \\ 
2-stage & 37.43 & 31.07 & 34.25 \\ 
\arrayrulecolor{black}\bottomrule
\specialrule{0em}{0.3em}{0em}
\multicolumn{4}{c}{{B. PSNRs of pre-training variants.}}\\
\end{tabular} 
\\

\end{tabular}
\label{tab:ablation}
\vspace{-1.5em}
\end{table}

\subsection{Ablation Studies}
\label{sec:ablation}

We conduct extensive ablation experiments to validate the effectiveness of the proposed CSformer model and the MAEIP pre-training method.
We first evaluate the effectiveness of the CSformer architecture on the SIDD dataset \cite{sidd} using the variant CSformer-T2.
We also design several pre-training variants and validate the effect of the MAEIP framework on image deraining \cite{mspfn2020} using CSformer-T.

\paragraph{Improvements on CSformer architecture.} The number of blocks for each hierarchical stage in CSformer-T2 is same with the Uformer-T \cite{wang2021uformer}. 
The second row in Table \ref{tab:ablation}-A shows the performance of Uformer-T on SIDD \cite{sidd} test set.
By replacing the LeFF module with the GCFFN (GC) module, the performance achieves 0.02 dB gain while reducing the computational complexity.
The original Uformer-T is an asymmetric structure, of which the number of channels in the decoder is twice the number of channels in the corresponding encoder.
By adjusting this asymmetric structure to the symmetric structure (Sym), in which the corresponding encoder and decoder have the same number of channels, the model further achieves a 0.04 dB gain.
Moreover, by adopting the global MSA (GA) in the bottleneck stage, the performance can be further improved.
Finally, we add channel-attention modules (CA) to the model and adjust the mlp ratio to achieve a similar computational complexity, which further improves the performance by 0.03 dB.
Through these improvements, our CSformer-T2 achieves the same performance as Uformer-S (39.77 dB) with fewer computational costs (13.50 GMACs \textit{vs.} 43.86 GMACs).

\paragraph{Effectiveness of the MAEIP pre-training.} We consider five different pre-training variants including (1) without pre-training (w/o p.), (2) only reconstructing encoder output (r. encoder), \textit{i.e.}, only pre-training encoder, (3) only reconstructing decoder output (r. decoder), (4) our MAEIP framework, \textit{i.e.}, reconstructing encoder output and decoder output simultaneously, and (5) a 2-stage pre-training.
We conduct this ablation experiment based on CSformer-T, which is a more lightweight model, and pre-train these variants for 60 epochs.
Table \ref{tab:ablation}-B demonstrates the finetuned performance of these pre-training variants on the image deraining task.
We follow the benchmark experimental settings and report the test results on the Rain100L \cite{yang2017deep} and Test100 \cite{zhang2019image} datasets.
Our MAEIP pre-training boosts the performance of CSformer-T by 0.5 dB on average, which is better than only reconstructing the encoder output or only reconstructing the decoder output.
For 2-stage pre-training, we first pre-train the encoder for 30 epochs, then pre-train the whole framework for 30 epochs.
The 2-stage pre-training achieves similar performance but the first-stage pre-training saves about half of the training time (25\% time saving overall), which is a more efficient pre-training method.
Moreover, as shown in Table \ref{table:denoising_real}, our MAEIP framework still works well for large CSformer models.

\begin{table}[t]
    \centering
    \setlength{\tabcolsep}{0.3em}
    \caption{PSNR results of different pre-training length.}
    \vspace{0.3em}
    \vspace{1pt}
    \label{tab:ablation2}
    \scalebox{0.8}{
    \begin{tabular}{l|x{60}x{55}x{55}}
    \toprule
        &  Rain100L \cite{yang2017deep} & Test100 \cite{zhang2019image} & Average \\
    \midrule
    w/o pre-training &  37.03    & 30.53   & 33.78 \\
    30 epochs        &  37.28    & 30.95   & 34.11 \\
    60 epochs        &  37.48    & 31.07   & 34.28 \\
    \bottomrule
    \end{tabular}
    }
    \vspace{-1.3em}
\end{table}

\paragraph{Influence of the pre-training length.} We also validate whether the training schedule length influences the finetuned performance.
As shown in Table \ref{tab:ablation2}, the CSformer-T pre-trained for 30 epochs significantly performs better than without pre-training, but worse than pre-trained for 60 epochs, which shows that the performance improves steadily with longer pre-training \cite{he2022mae}.

\section{Conclusion}
\vspace{-0.5em}
This paper demonstrates that masked autoencoders are also effective self-supervised learning frameworks for image processing tasks.
We first devise an efficient CSformer model, which considers both channel attention and shifted-window-based self-attention.
Then we propose an effective MAE architecture for image processing tasks termed MAEIP, and implement the pre-training process based on the proposed excellent CSformer model.
Extensive experiments show that the CSformer pre-trained with the self-supervised framework MAEIP achieves state-of-the-art performance on several image processing tasks, and our CSformer is also computationally efficient.
We demonstrate a few applications, but our pre-trained CSformer is also expected to perform well on other tasks and applications.

{\small
\bibliographystyle{ieee_fullname}
\bibliography{bib}
}

\end{document}